\documentclass[twoside,11pt]{article}

\usepackage{jmlr2e}
\usepackage{algorithm}
\usepackage{algorithmic}

\usepackage{amsmath}
\usepackage{amsfonts}

\usepackage{xcolor}

\usepackage{adjustbox}

\usepackage{rotating,booktabs}
\usepackage[a4paper,margin=2.5cm]{geometry}

\usepackage{float}
\restylefloat{table}

\usepackage{subcaption}

\newcommand{\ie}{i.e.}

\ShortHeadings{Rule-Mining based classification}{Luck M., Pallet N., Damon C.}
\firstpageno{1}

\begin{document}

\title{Rule-Mining based classification: a benchmark study}

\author{\name Margaux Luck \email margaux.luck@gmail.com \\
       \addr University Paris Descartes and Institut Hypercube\\
       Paris, France \\
       \AND
       \name Nicolas Pallet \email nicolas.pallet@parisdescartes.fr  \\
       \addr University Paris Descartes \\
       Paris, France \\
       \AND
       \name Cecilia Damon \email cecilia.damon@gmail.com \\
       \addr Institut Hypercube\\
       Puteaux, 92907, France}

\editor{}

\maketitle

\begin{abstract}%   <- trailing '%' for backward compatibility of .sty file
This study proposed an exhaustive stable/reproducible rule-mining algorithm combined to a classifier to generate both accurate and interpretable models. Our method first extracts rules (\ie, a conjunction of conditions about the values of a small number of input features) with our exhaustive rule-mining algorithm, then constructs a new feature space based on the most relevant rules called "local features" and finally, builds a local predictive model by training a standard classifier on the new local feature space. This local feature space is easy interpretable by providing a human-understandable explanation under the explicit form of rules. Furthermore, our local predictive approach is as powerful as global classical ones like logistic regression (\textit{LR}), support vector machine (\textit{SVM}) and rules based methods like random forest (\textit{RF}) and gradient boosted tree (\textit{GBT}).
\end{abstract}


\section{Introduction}
The most popular machine learning methods such as logistic regression \textit{LR}) [\cite{hosmer2013applied}], support vector machines (\textit{SVMs}) [\cite{cristianini2000introduction}] or neural networks [\cite{goodfellow2016deep}] usually perform quite well classification tasks. However, these methods present at least three disadvantages. First, they often works as a black box generating models not easily understandable and interpretable [\cite{tickle1998truth,barakat2010rule}]. Second, many machine learning methods do not explore the local and intense phenomena which are inherent to the data nature and complexity. Third, they make the strong assumption that the classification boundary is contiguous. For example, in biological processes, local phenomena generated by the heterogeneity of the populations, the complex composition of biological data and huge local interactions between features may play a major role [\cite{holmes2002chemometric,romero2016diet}]. Also, more and more studies include heterogeneous data increasing the amount of interaction between features and therefore the presence of local signal of interest.

Consequently, the use of a classification rule learning approach such as decision-tree based methods like random forest (\textit{RF}) [\cite{breiman2001random}] and gradient boosted tree (\textit{GBT}) [\cite{drucker1995boosting}] appears to be a valuable approach to discover local phenomena within feature subspaces and notably to provide a human-understandable explanations under the explicit form of rules. Moreover, contrary to the off-the-shelf classifiers, they are able to handle data as complex as missing values, numerical and categorical data, multi-collinear features, outliers and local relationships among features. However, these ensemble methods only explore random parts of the original feature space and are again difficult to interpret ("black box"). Indeed, as an example, \textit{RF} operate by building a forest of uncorrelated trees, each representing a random subspace of the whole feature space, which are combined by bagging approach to generate the final decision model [\cite{breiman2001random}]. Moreover, classification rule learning methods are limited in their subgroup discovery task since they are looking for the best set of rules predicting all the classes of the target features.

Another alternative is subgroup discovery methods that aim at extracting individual rules of interest [\cite{herrera2011overview}]. Standard classification- and association- rule learning methods have been adapted to subgroup discovery task [\cite{kavvsek2006apriori}]. Concretely, an algorithm of subgroup discovery will look for subsamples characterized by a subset of some features (rules) in which the phenomenon is very intense. It is critical to understand that the subgroup discovery task may lack classification power compared to the classification task. Indeed, the subgroup discovery task focuses on discovering discriminative knowledge (\ie, relevant population subgroups) with a covering strategy less specific than the classification task focusing on maximizing classification accuracy [\cite{cano2008subgroup}]. 

Some studies decided to combine the benefits of both tasks in order to learn classification models which are as efficient as they are interpretable [\cite{karabatak2009expert, kianmehr2008carsvm}]. There are many other arguments in favor of this approach: ability to 1) describe unusual interesting distributions, 2) effectively solve contiguous or non-contiguous classification problems with discrete boundaries, 3) analyze heterogeneous, complex and missing data [\cite{friedman2008predictive,tang2005granular}]. As the exhaustive search for the best subgroups results in a high combinatorial complexity, most subgroup discovery algorithms used heuristic search [\cite{lavravc2004subgroup}]. The major drawback of an heuristic search is that it could miss good subgroups and lead to over-complex and locally optimal solutions which do not generalize well to unseen data. As described in [\cite{luck2016rule}], we proposed an exhaustive stable/reproducible rule-mining combined to a classifier to generate both accurate and interpretable models. Our method first extracts rules (\ie, a conjunction of conditions about the values of a small number of input features) with our exhaustive rule-mining algorithm, then constructs a new feature space based on the most relevant rules called "local features" and finally, builds a local predictive model by training a standard classifier on the new local feature space.

In this study, we went further in the evaluation of the rules' quality, the definition of the local features and the optimization of the rule-mining algorithm. A comprehensive benchmark study has been conducted to compare our method with a set of classification methods on synthetic and real datasets through three levels of comparison.

\section{Local predictive models}
\label{sec:rulealgo}
In this section, we describe the different parts for the construction of the three different strategies of local predictive modelisation. They first extracted local features based on their own rule-mining algorithm and then built a local predictive model on these local features with off-the-shelf classifiers.

\subsection{Rule-mining strategies}
\textbf{Our rule-mining algorithm.} A complete description of our rule-mining algorithm is detailed in [\cite{luck2016rule}] and an open-source implementation is available on git-hub (\url{https://github.com/Museau/Rule-Mining.git}). The rule-mining strategy of our algorithm (see Algorithm 1:step 1) consists of an exhaustive exploration of the multi-dimensional feature space in order to extract regions (\ie, hypercube) having an over-concentration of a class of subjects. We recall that these regions may be represented under the form of a rule:
\[\text{Rule: } \{\text{Feature(s) condition(s)}\}\]
where feature(s) condition(s) correspond to one value or a range/set of values of a given feature observed in the dataset.
\\

\textbf{Association rules based rule-mining.} Association rules [\cite{borgelt2012frequent}] describe relationships between features in a dataset based on the search of frequent item sets. 
Let $I = {i_{1}, i_{2}, ..., i_{n}}$ be a set of items, \ie, all the unique values of the features, and $T = \{t_{1}, t_{2}, ..., t_{m}\}$ be a set of transactions or samples, such that $t_{i}$ is a subset of $I$ (\ie, $t_{i} \subseteq I$). An association rule is represented under the form: 
\[X \rightarrow Y \text{, where } X \in T \text{, } Y \in T \text{, and } X \cap Y = \emptyset\]
where X is a subset of I called item-set. In our case, we are looking for all the association rules which induce one of the different classes of samples (\ie, target feature). Therefore, we constrained the association rule induction to select only rules having as a consequent $Y$ one of the values assigned to the target features.\\

\textbf{Decision tree based rule-mining.} A decision tree [\cite{breiman1984classification}] is a tree-like structure, which starts from root features, and ends with leaf nodes. Generally a decision tree has several branches consisting of different features conditions (\ie, threshold on feature values). The leaf node on each branch is assigned a class label as the majority class of the samples in its parent node. Here, the tree was generated according to the Gini impurity measure. For each leaf, we extracted a rule as being the set of nodes (\ie, decisions) from the root to the leaf.

\subsection{Rule selection}
The rule selection step in \textbf{our rule-mining algorithm} proceeds in two stages.
\begin{itemize}
\item[1)] It computes two measures of rule quality for each class of samples and selects rules on the basis of these two measures (see Algorithm~\ref{algo}:Step 2.a): 

\begin{itemize}
\item[-] rules with a z-score greater to 1.96 corresponding to a significant difference between the proportion of the outcome class in the rule (\(p\)) and in the total population (\(p_0\)) (with a confidence level of 95\%). The z-score is defined as follows:
\[\text{z-score} =  \sqrt{n} \frac{p-p_0}{\sqrt{p(1-p_0)}}\]
where n is the number of subjects.
\item[-] rules with a class size above a threshold fixed according to the following formula: 
\[\text{number of samples} / \text{number of bins} * \% \text{ class samples}\]
\end{itemize}

\item[2)] We reduced the set of previously selected rules by removing some redundant rules, \ie, those sharing the same feature(s) and overlapping condition(s) but having worse quality measure (see Algorithm~\ref{algo}:Step 2.b and 2.c). The objective is to keep, for continuous features only both disjoint and discriminative rules, and, for discrete features only the most discriminative and covering rules.
\\
\end{itemize}

For \textbf{association rules based rule-mining}, we selected rules with the stage 1 of the rule selection process used for our algorithm. \\

Finally, for \textbf{decision tree based rule-mining}, we selected the rules by fixing the minimum size of a leaf equal to: number of samples$/$number of bins $*$ $\%$  minority class samples.

\begin{algorithm}
\caption{\textbf{Supervised rule mining}}
\textbf{Input:} \\
- features\\
- target\\
- rule dimension\\
- the thresholds of the 2 quality measures (z-score and rule size)\\
\textbf{Output:}\\  
- a set of relevant rules for each class\\

For rule dimension $=$ 1 to d:
\begin{quote}
Construct all couples (feature(s), target) with Card(features(s)) $=$ d
\end{quote}
\noindent\rule{0.4\textwidth}{0.4pt}\\
\textbf{Step 1: Exhaustive  rule generation}\\
Input:\\
- a couple (feature(s), target)\\
Output:\\
- a set of candidate rules for each modality\\

For feature in feature(s)
\begin{quote}
If discretized continued feature
\begin{quote}
Construct all possible candidate rule combinations made from adjacent bins.
\end{quote}
If discretized categorical feature
\begin{quote}
Construct all possible candidate rule combinations made from the Cartesian product of the bins with no repeated elements
\end{quote}
\end{quote}
If length(feature(s))$>$1:
\begin{quote}
Construct all possible multivariate candidate rule combinations based on the previous univariate candidate rules
\end{quote}

\noindent\rule{0.4\textwidth}{0.4pt}\\
\textbf{Step 2: Relevant rule selection}\\
Input:\\
- a set of candidate rules\\
Output:\\
- a set of significant rules\\

a.For each target class
\begin{quote}
keep the rules with a z-score and rule size \(\geq\) fixed thresholds
\end{quote}
b. For each target class
\begin{quote}
Delete rules that are included in bigger rules with smaller z-score.
\end{quote}
\begin{quote}
c. If discretized continued feature:
\begin{quote}
Delete rules covering smaller rules with higher z-score.
\end{quote}
\end{quote}
\label{algo}
\end{algorithm}

\subsection{Local feature description}
To construct our local predictive models, we need to use the rules extracted with the different rule-mining strategies as new input to the learning process of the classifiers.\\\\
For \textbf{our rule-mining algorithm}, we carried each rule information (\ie, features conditions) over to all the training samples yielding to a vector of length number of training samples, called local feature. We proceeded in three ways according to the features types: 
\begin{itemize}
\item[a)] Rules that have been generated based on discrete features are represented as a binary vector where a value of 1 (resp. 0) is assigned to all the samples inside (resp. outside, \ie, samples that do not check the discrete features conditions) the rule; 

\item[b)] For rules generated based on continuous features, we developed a distance measure, called here $\Delta$ and inspired by the work of \cite{salzberg1991nearest}. $\Delta$ represents the distance of a sample to the rule center and is defined as:

\[\Delta = w_{r} * \sqrt{ \sum_{i=1}^{m} (w_{i} * \delta_{i})^{2}}\]

where $w_{r}$ is the weight associated to the rule $r$ (here, the z-score), $w_{i}$ the weight associated to the feature $i$ (here, the feature frequency among the final set of rules) and $\delta_{i}$ the distance of the sample to the rule center considering only the feature condition on the feature $i$. $\delta_{i}$ is set to one for samples that are at the rule center. Otherwise, $\delta_{i}$ is defined as following:
\begin{equation*}
\begin{split} 
\delta_{i} = & 1 - ((x_{i} - r_{\text{center}})/(\max(X_{i})-\min(X_{i})) \\
 			& if x_{i} > r_{\text{center}} \\
\delta_{i} = & 1 - ((r_{\text{center}}-x_{i})/(\max(X_{i}) - \min(X_{i})) \\
 		    & if x_{i} < r_{\text{center}} \\
\text{with}\\
r_{\text{center}} = & r_{i_{\text{low}}} + ((r_{i_{\text{up}}} - r_{i_{\text{low}}})/2)
\end{split}
\end{equation*}

where $r_{i_{\text{low}}}$ $r_{i_{\text{up}}}$ are respectively the lower and upper limit of the rule and $r_{\text{center}}$ is the center of the rule
\item[c)] For rules generated based on mixed features, we assigned a value of $\Delta$ (resp. 0) to all samples that (resp. do not) check the discrete features conditions.  \\
\end{itemize}

For \textbf{association rules} and \textbf{decision tree based rule-mining}, we used the binary representation described above. This is because the association rules technique searches for frequent unique bin values and not ranges of bin values which corresponds to the discrete case of our rule-mining algorithm. For decision tree, all training samples are successively splitted at each node (from the root to the leaf) according to a binary decision, therefore belonging or not to the final global rules .  

\subsection{Classification}
For the predictive classification modeling step, we tested two algorithms: the \textit{LR} [\cite{hosmer2013applied}] with a penalty in $L_{1}$ or in $L_{2}$ and the SVM [\cite{cristianini2000introduction}] with a radial basis function (RBF) or a linear kernel. These algorithms were trained on the local feature matrix extracted from one or the other of the methods.

\section{Benchmark study}
In this section we will first present the three levels of comparison of our benchmark study. Then, we briefly describe one synthetic and five real-life datasets used to test the different learning methods and finally, the learning-test evaluation procedure.

\subsection{Three levels of comparison}
The major goal of this benchmark study is to prove the benefit of using local features, derived from our supervised rules extraction method, prior to the classification step. Regarding our predictive local method, detailed in Section~\ref{sec:rulealgo}, it consists of two steps: 1- the generation and selection of local features based on rules extracted with our rule-mining algorithm (see Algorithm~\ref{algo}) and 2- the classification step.

\begin{figure}[!htb]
\centering
\includegraphics[scale=0.45]{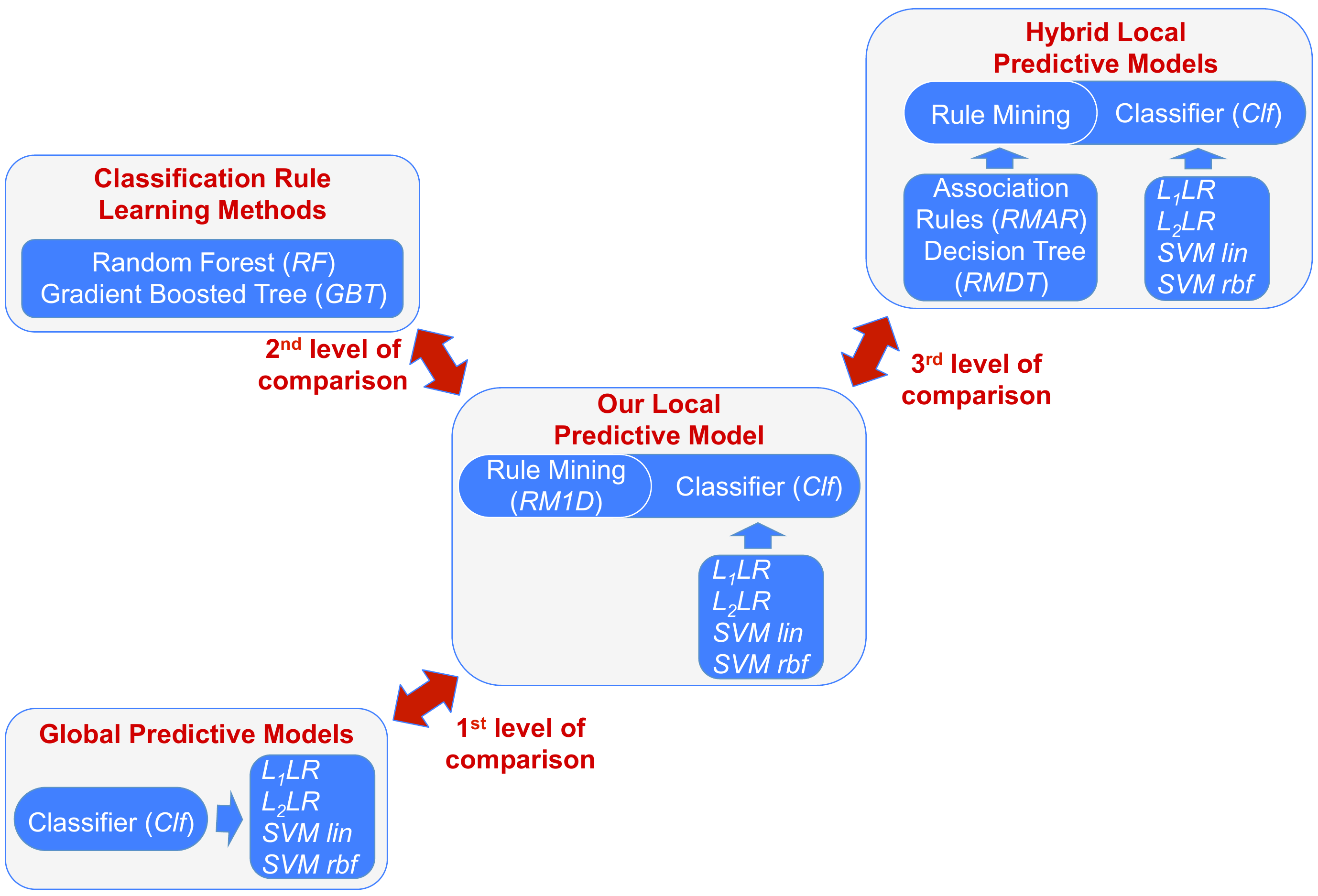}
\caption{The three levels of comparison used in the benchmark study.}
\label{benchmark_study}
\end{figure}

\begin{flushleft}In this benchmark study, we emphasized three levels of comparison (see Figure \ref{benchmark_study}):
\begin{description}
\item[1\textsuperscript{st} level:]comparison of our local predictive methods (\textit{RM1D-Clf}) with the classifiers (\ie, denoted by \textit{Clf}) used in our local predictive methods but applied in a classical way on the original feature space (\ie, global predictive models). For that, we choose to use a set of different and widely used classifiers: logistic regression with either a penalty in $L_{1}$ ($L_1LR$) or $L_{2}$ ($L_2LR$) [\cite{hosmer2013applied}] and support vector machines with linear (\textit{SVM lin}) or radial basis function (\textit{SVM rbf}) kernel [\cite{cristianini2000introduction}]. The goal of this first level of comparison is to prove the added-value of rules for prediction purpose especially in terms of interpretation; 

\item[2\textsuperscript{nd} level:]comparison of our local predictive method (\textit{RM1D-Clf}) with well-known classification rule learning methods: random forest (\textit{RF}) and gradient boosted tree (\textit{GBT}) [\cite{breiman2001random}]. The goal of this second level of comparison is to understand the interest of extracting individual discriminative rules rather than extracting the best set of rules in terms of prediction; 

\item[3\textsuperscript{rd} level:]comparison of our local predictive method (\textit{RM1D-Clf}) with two other hybrid local predictive methods using rules from decision tree (\textit{RMDT-Clf}) [\cite{breiman1984classification}] or association rules (\textit{RMAR-Clf}) techniques [\cite{agrawal1993mining}] as input data to the classifiers. The goal of this third level of comparison is to assess different strategies of rules' extraction.
\end{description}
\end{flushleft}
Publicly available implementations of all the known methods used in the benchmark come from the Scikit-learn project [\cite{pedregosa2011scikit}] for logistic regression (\textit{LR}), support vector machines (\textit{SVMs}), random forest (\textit{RF}), gradient boosted tree (\textit{GBT}) and decision tree (\textit{DT}), and pyfim module [\cite{borgelt2012frequent}] for association rules (\textit{AR}).

\subsection{Datasets}
We assess all the comparative experiments of our benchmark study on five real-life datasets and one synthetic dataset. The characteristics of these datasets are resumed in table~\ref{table_datasets}.

\subsubsection{Real-life datasets}
We selected 5 real-life datasets representing different real life problematics. As you can see in Table~\ref{table_datasets}, these datasets have different characteristics in terms of number of classes, features' type (\ie, continuous, discrete or mixed) and dimension (\ie, number of instances and number of features) and presence of missing values. Below, we briefly describe each of these real-life dataset obtained from the UCI Machine Learning Repository [\cite{merz1996uci}].

\textbf{The Wisconsin Breast Cancer (Diagnostic) or WDBC dataset} is based on 10 real-values features from the cell nuclei observed in breast images. The mean, the standard error and the worst (mean of the three largest values) of each feature were computed for the dataset, resulting in a total of 30 features [\cite{street1993nuclear}].
The learning task is to classify each example as malignant (cancerous) or benign (non-cancerous). The number of malignant and benign are 212 (37 $\%$) and 357 (63 $\%$), respectively.

\textbf{The wine dataset} records the measurements of 13 chemical constituents (e.g., \textit{alcohol}, \textit{flavanoids}) of the wines. The learning task is to classify the 178 wines' samples into 3 cultivators from Italy. The number of examples by cultivators is respectively 48 (27 $\%$), 59 (33 $\%$) and 71 (40 $\%$).

\textbf{The iris dataset} extracts 4 features of the iris plants. The learning task is to classify the 150 samples into 3 categories of iris (iris-setosa, iris-versicolour, or iris-virginica). The number of samples in each class is exactly one third of the samples (\ie, 50 examples).

\textbf{The balance scale dataset} presents the results of some psychological experiments which are a balance scale tip to the right (288 instances, \ie, 46 $\%$), tip to the left (288 instances, \ie, 46 $\%$), or be balanced (49 instances, \ie, 8 $\%$). The correct way to find the class is based on 4 discrete features, which are the \textit{left}/\textit{right weights} and \textit{distances}, according to the following formula: class $=$ tip to the left (resp. right, balanced) if (\textit{left distance} $*$ \textit{left weight}) $>$ (resp. $<$, $=$) (\textit{right distance} $*$ \textit{right weight}).

\textbf{The Cleveland heart disease} dataset contains some clinical and noninvasive test results (\ie, 13 continuous and discrete features) of 303 patients. The goal is to identify the patients with heart disease (139 of 303 patients, \ie 46 $\%$).

\begin{table}[!ht]
\centering
\small
\begin{tabular}{ |l|c|c|c|c|c| } 
\hline
Datasets & N class & Features' type & N instances & N features & Missing values\\
\hline
WDBC & 2 & continuous & 569 & 30 & no \\
Wine & 3 & continuous & 178 & 13 & no \\
Iris & 3 & continuous & 150 & 4 & no \\
Balance scale & 3 & discrete & 625 & 4 & yes \\
Heart disease & 2 & mixed & 303 & 13 & yes \\
Synthetic & 3 & mixed & 500 & 4 & no \\
Synthetic noisy & 3 & mixed & 500 & 4 & no \\
\hline
\end{tabular}
\caption{\textbf{Characteristics of the real-life and synthetic datasets.}}
\label{table_datasets}
\end{table}

\subsubsection{Synthetic dataset}
We constructed artificial data defined by four features ($x_{1}$, $x_{2}$, $x_{3}$, $x_{4}$), two of which are continuous ($x_{1}$; $x_{2} \sim \mathcal{U}[0, 1]$) and the two other are discrete ($x_{3} = \{0, 1\}$; $x_{4} = \{\text{blue, white, red}\}$).

Each sample was labeled with one of the three classes $y=\{0, 1, 2\}$ according to the following rules: 
\begin{align}
  y=
  \begin{cases}
    0  & \text{if }  
    \begin{array}{l}
		{x_4=\text{red}, x_{3}=1}\\
        {x_4=\text{red}, x_{3}=0, x_{2} <= 0.5}\\
        {x_{4}=\text{blue or white}, x_{1} >= 0.7, x_{3}= 0, x_{2} > 0.2}\\
        {x_{4}=\text{white}, x_{1} \in ]0.5, 0.7[}\\
        {x_{4}=\text{blue}, x_{1}<=0.7}\\\\
    \end{array}\\
    1 & \text{if }  
    \begin{array}{l}
    	{x_{4}=\text{red}, x_{3} = 0, x_{2} > 0.5}\\
        {x_{4}= \text{blue or white}, x_{1} >= 0.7, x_{3}=0, x_{2} <= 0.2}\\
        {x_{4}=\text{white}, x_{1} <= 0.5}\\\\
	\end{array}\\
    2 & \text{if }  {x_{4}=\text{blue or white}, x_{1} >= 0.7, x_{3} = 1}\\
  \end{cases}
  \label{synthetic_dataset}
\end{align}

\begin{figure}[!htb]
\centering
\includegraphics[scale=0.45]{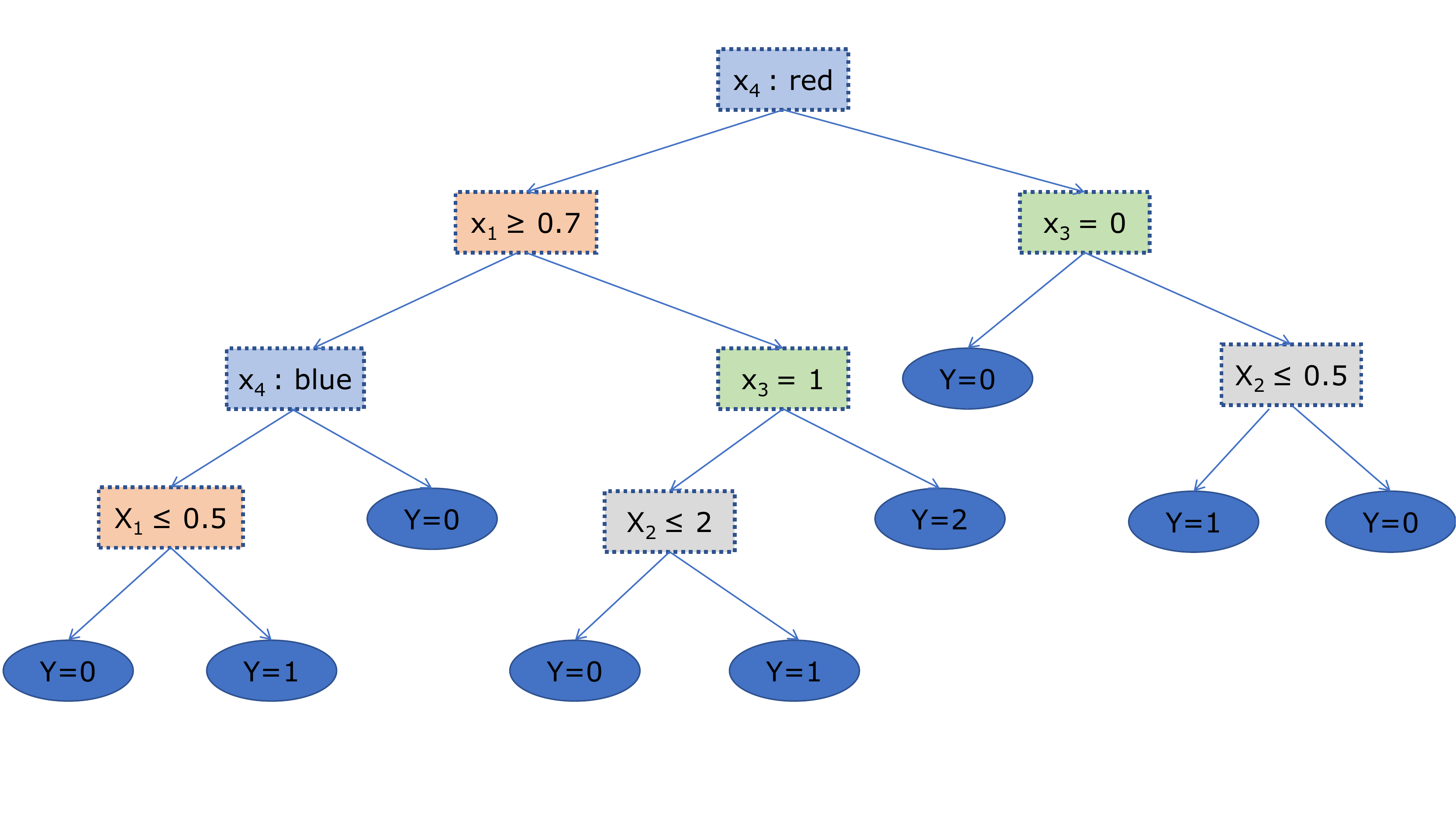}
\caption{Decision tree used to generate synthetic data and defining the rules' system (see equation \ref{synthetic_dataset}).}
\label{syntheticdataset}
\end{figure}

These rules can be represented under the form of a decision tree (see Figure \ref{syntheticdataset}). Based on these artificial data, we generated two synthetic datasets of 500 samples, one without noise and the other included approximately 12 $\%$ of noise. Noise reflects a decrease of rules purity through the introduction of classification errors within each rule. Each one of the three classes ($y=\{0, 1, 2\}$) includes respectively 340, 123 and 37 samples.

\subsection{Data preprocessing}
For methodological comparison purpose, for a given dataset, we decided to apply the same data preprocessing prior to all learning strategies (even if it is not necessary for some strategies, particularly for the global models). Therefore, despite the fact that the \textit{RMAR} and \textit{RM1D} algorithms operates with or exploits missing values, we first completed missing values in order to avoid the removal of numerous samples from the different affected datasets necessary for the use of the other classifiers. This data filling is done with the most frequent category for discrete features and the median value for continuous features. We also discretized the continuous features with a 10 bins quantization, in order to put the features in the same scale. Moreover, this discretization step is a prerequisite to our local predictive models and to the association rule algorithm. This step is necessary to avoid obtaining rules that are to small (in some cases consisting of a single subject) especially in the case of association rules. With a 10 bins quantization, we constrained the minimum size of a rule to 10 $\%$ of the total number of samples for each dataset. For discrete features, we used one hot encoding except for \textit{AR} and our rule-mining algorithm were we keep the categories as they were. These choice was made depending of the implementation of the different algorithms. We have studied the impact of this discretization and data filling on the performances of our different learning strategies. 

\subsection{Evaluation procedure}
We first split the datasets into train (70 $\%$) and test (30 $\%$) sets in which the percentage of samples for each class is preserved. We performed this step 5 times to avoid being dependent on a particular random selection of a test set, which could lead to biased results. More importantly, it allowed us to assess the stability and robustness of the different learning strategies. 

For each of these 5 splits, we optimized the classifier parameters with an 5-fold inner cross-validation on the training set. For \textit{SVM lin}, \textit{SVM rbf}, $L_{1}LR$ and $L_{2}LR$ classifiers, we optimized the regularization parameter $C \in [1e^{-3}, 1e^{-2}, 1e^{-1}, 1e^0, 1e^1, 1e^2, 1e^3]$. For \textit{RF} and \textit{GBT} classifiers, we optimized the number of trees in the forest from the list of numbers $[100, 200, 300, 400, 500]$. For \textit{GBT}, we also optimized the learning rate within the range $[1e^{-3},1e^{-2}, 1e^{-1}, 1e^0, 1e^1, 1e^2]$. Concerning the three rule mining methods, \textit{RM1D}, \textit{RMAR} and \textit{RMDT}, we set a priori some parameters, specifically the size of the final rules defined according to the datasets', classes' and bins' sizes. For \textit{RM1D} and \textit{RMAR}, we also defined a standard statistical threshold on a measure of over-representation of a class within the rules.

As performance scores, we computed the average and standard deviation of the F1 scores over the 5 splits. It is important to note that the classification errors are weighted in inverse proportion to class frequencies in the input data, and multi-class problems are solved with a one-vs-rest decision function.

In addition to the predictive power, we assessed the stability of the local models over the 5 splits. Local common features measures were computed using the Jaccard similarity coefficient, which measures the similarity between two finite sets of selected features used for the prediction of two models: 
\[J = \frac{|A \bigcap B|}{|A \bigcup B|}\] where $|A \bigcap B|$ and $|A \bigcup B|$ are the intersection and union size of the two sets $A$ and $B$. Then, we reported the mean and standard deviation of the two-by-two Jaccard coefficients computed over the 5 splits. For the union and intersection of two sets of local features A and B, we considered all the tuples (feature, single value) contained in the local features. We computed the union and intersection of the Jaccard index over all these tuples.  

\section{Results and discussion}
\subsection{General consideration}
The correspondences between the bins resulting from the quantization process and the original values for the WDBC, wine, iris, heart disease, synthetic and synthetic noisy datasets are reported in table~\ref{bin_wdbc} to~\ref{bin_synthetic} respectively in appendix A. It is important to note that the quantization process did not have a significant impact on the performances of the global linear classifiers, \ie, $L_{1}LR$, $L_{2}LR$, \textit{SVM lin}, leading to generally similar results. However, for the global non-linear classifier \textit{SVM rbf}, we observed higher (resp. lower) performances for the WDBC, wine, synthetic and synthetic noisy (resp. iris and heart Disease) datasets with the quantization process than without. All these differences do not alter the following conclusions. We also observed that the completion of missing values has no effect on the performances of \textit{RMAR} or \textit{RM1D} which usually function well with missing values for the two datasets concerned.

It should also be noted that we tested both one-dimensional (1D) and two-dimensional (2D) rules with our rule-mining algorithm (\ie, \textit{RM1D} and \textit{RM2D}). We did not report the results obtained with 2D rules because they have no added-value in term of performance compared to those obtained with 1D rules. Moreover, the resulting models were much more complex including redundant information, more difficult to interpret and required heavy computation time.

In the following subsections, we compared the different methods through all the datasets according to the three levels of comparison, in terms of predictive performance, complexity and stability. Then, we examined in more detail the rules obtained on each dataset with our local predictive method.   

\subsection{1st level of comparison: "Global \textit{vs} our Local Predictive Models"}
\begin{sidewaystable}
\centering
\small
\resizebox{\textwidth}{!}{%
\begin{tabular}{@{} l *{9}{c} @{}}
\toprule
\textbf{Benchmark} & \multicolumn{7}{c@{}}{\textbf{Datasets}}\\
\cmidrule(l){2-8}
& \textbf{WDBC} & \textbf{Wine} & \textbf{Iris} & \textbf{Balance scale} & \textbf{Heart disease} & \textbf{Synthetic} & \textbf{Synthetic noisy} \\
\midrule
$L_{2}LR$ & 95.6 $\pm$ 2.1 (46.0) & 96.0 $\pm$ 2.4 (40.0) & 94.2 $\pm$ 2.6 (36.2) & 94.1 $\pm$ 7.2 (13.1) & \textcolor{red}{83.3} $\pm$ 1.1 \textbf{(75.7)} & 85.3 $\pm$ 2.0 (42.6) & 56.8 $\pm$ 3.1 (18.3) \\
\textit{RM1D-$L_{2}LR$} & \textcolor{red}{97.0} $\pm$ 0.7 \textbf{(138.6)} & 98.1 $\pm$ 1.9 (51.6) & 97.3 $\pm$ 0.9 (108.1) & 95.9 $\pm$ 4.7 (20.4) & 83.1 $\pm$ 1.7 (48.9) & 84.1 $\pm$ 2.2 (38.2) & 60.8 $\pm$ 4.2 (14.5) \\
 \addlinespace
$L_{1}LR$ & 95.9 $\pm$ 1.7 (56.4) & 97.6 $\pm$ 1.4 (69.7) & 95.1 $\pm$ 2.6 (36.6) & 97.4 $\pm$ 2.9 (33.6) & 82.7 $\pm$ 1.3 (63.6) & 85.3 $\pm$ 2.0 (42.6) & 56.0 $\pm$ 3.3 (17.0) \\
\textit{RM1D-$L_{1}LR$} & 96.2 $\pm$ 1.3 (74.0) & 97.0 $\pm$ 1.4 (69.3) & 97.3 $\pm$ 0.9 (108.1) & 97.4 $\pm$ 2.9 (33.6) & 83.1 $\pm$ 1.5 (55.4) & 83.3 $\pm$ 2.3 (36.2) & 58.6 $\pm$ 3.7 (15.8) \\
 \addlinespace
\textit{SVM lin} & 96.9 $\pm$ 1.3 (74.5) & 98.0 $\pm$ 1.8 (54.4) & \textcolor{red}{98.2} $\pm$ 0.9 \textbf{(109.1)} & \textcolor{red}{98.2} $\pm$ 1.1 (89.3) & 82.8 $\pm$ 1.8 (46.0) & 88.0 $\pm$ 2.6 (33.8) & 60.5 $\pm$ 6.5 (9.3) \\
\textit{RM1D-SVM lin} & 96.4 $\pm$ 0.9 (107.1) & 98.0 $\pm$ 2.6 (37.7) & \textcolor{red}{98.2} $\pm$ 0.9 \textbf{(109.1)} & 97.9 $\pm$ 0.6 \textbf{(163.2)} & 82.7 $\pm$ 2.2 (37.6) & 88.0 $\pm$ 0.8 \textbf{(110.0)} & 69.2 $\pm$ 4.5 (15.4) \\
 \addlinespace
\textit{SVM rbf} & 88.7 $\pm$ 2.5 (35.5) & 81.5 $\pm$ 5.6 (14.6) & \textcolor{red}{98.2} $\pm$ 0.9 \textbf{(109.1)} & 85.8 $\pm$ 5.0 (17.2) & 76.5 $\pm$ 4.4 (17.4) & 92.4 $\pm$ 1.3 (71.1) & 71.6 $\pm$ 4.2 (17.0) \\
\textit{RM1D-SVM rbf} & 96.2 $\pm$ 1.0 (96.2) & \textcolor{red}{98.7} $\pm$ 1.2 \textbf{(82.2)} & \textcolor{red}{98.2} $\pm$ 0.9 \textbf{(109.1)} & 90.4 $\pm$ 3.2 (28.2) & 80.2 $\pm$ 4.1 (19.6) & \textcolor{red}{95.6} $\pm$ 1.2 (79.7) & \textcolor{red}{74.2} $\pm$ 3.9 \textbf{(19.0)} \\
 \addlinespace
 \textit{Clf var.} & 10.6 & 46.8 & 3.3 & 24.1 & 7.8 & \textbf{8.4} & \textbf{38.8} \\
\textit{RM1D-Clf var.} & \textbf{0.1} & \textbf{0.4} & \textbf{0.2} & \textbf{8.9} & \textbf{1.3} & 23.7 & 39.7 \\
\addlinespace
\bottomrule
\end{tabular}}
\caption{\textbf{Performance scores of the models of the 1st level of comparison.} The table reports the F1 score mean$\pm$sd (mean/std) over the 5 splits. For each dataset, the best model regarding the mean (resp. mean/std) score is highlighted in red (resp. bold). The values listed at the bottom of the table are the variances of the mean F1 scores over the global and local models respectively (\ie, \textit{Clf} and \textit{RM1D-Clf var.}). }
\label{results_F1score_1}
%\end{table}
\resizebox{\textwidth}{!}{%
\begin{tabular}{@{} l *{9}{c} @{}}
\toprule
\textbf{Benchmark} & \multicolumn{7}{c@{}}{\textbf{Datasets}}\\
\cmidrule(l){2-8}
& \textbf{WDBC} & \textbf{Wine} & \textbf{Iris} & \textbf{Balance scale} & \textbf{Heart disease} & \textbf{Synthetic} & \textbf{Synthetic noisy} \\
\midrule
\textit{RF} & 95.9 $\pm$ 2.2 (43.6) & \textcolor{red}{99.0} $\pm$ 1.3 (76.2) & 96.9 $\pm$ 1.1 (88.1) & 49.6 $\pm$ 0.5 (99.2) & 82.1 $\pm$ 2.2 (37.3) & 94.0 $\pm$ 1.3 (72.3) & 72.0 $\pm$ 6.3 (11.4) \\
\textit{GBT} & 96.5 $\pm$ 1.6 (60.3) & 92.1 $\pm$ 4.9 (18.8) & 95.5 $\pm$ 2.0 (47.7) & 53.1 $\pm$ 4.2 (12.6) & 81.0 $\pm$ 3.4 (23.8) & 94.0 $\pm$ 1.3 (72.3) & \textcolor{red}{75.7} $\pm$ 5.3 (14.3) \\
 \addlinespace
\textit{RM1D-Clf*} & \textcolor{red}{97.0} $\pm$ 0.7 \textbf{(138.6)} & 98.7 $\pm$ 1.2 \textbf{(82.2)} & \textcolor{red}{98.2} $\pm$ 0.9 \textbf{(109.1)} & \textcolor{red}{97.9} $\pm$ 0.6 \textbf{(163.2)} & \textcolor{red}{83.1} $\pm$ 1.5 \textbf{(55.4)} & \textcolor{red}{95.6} $\pm$ 1.2 \textbf{(79.7)} & 74.2 $\pm$ 3.9 \textbf{(19.0)} \\
\addlinespace
\bottomrule
\end{tabular}}
\caption{\textbf{Performance scores of the models of the 2nd level of comparison.} The table reports the F1 score mean$\pm$sd (mean/std) over the 5 splits. For each dataset, the best model in terms of mean (resp. mean/std) score is highlighted in red (resp. bold).}
\label{results_F1score_2}
\end{sidewaystable}

\begin{sidewaystable}
\centering
\small
\resizebox{\textwidth}{!}{%
\begin{tabular}{@{} l *{9}{c} @{}}
\toprule
\textbf{Benchmark} & \multicolumn{7}{c@{}}{\textbf{Datasets}}\\
\cmidrule(l){2-8}
& \textbf{WDBC} & \textbf{Wine} & \textbf{Iris} & \textbf{Balance scale} & \textbf{Heart disease} & \textbf{Synthetic} & \textbf{Synthetic noisy} \\
\midrule
\textit{RMDT-$L_{2}LR$} & 90.8 $\pm$ 3.3 (27.5) & 83.6 $\pm$ 4.1 (20.4) & 90.1 $\pm$ 4.2 (21.5) & 60.7 $\pm$ 3.5 (17.3) & 76.6 $\pm$ 2.7 (28.4) & 94.0 $\pm$ 1.4 (67.1) & 76.3 $\pm$ 5.8 (13.2) \\
\textit{RMAR-$L_{2}LR$} & 95.4 $\pm$ 1.8 (53.0) & 96.6 $\pm$ 2.9 (33.3) & 94.7 $\pm$ 2.2 (43.0) & 60.3 $\pm$ 4.0 (15.1) & 84.1 $\pm$ 1.5 \textbf{(56.1)} & 91.0 $\pm$ 1.1 (82.7) & 69.0 $\pm$ 4.7 (14.7) \\
\textit{RM1D-$L_{2}LR$} & \textcolor{red}{97.0} $\pm$ 0.7 \textbf{(138.6)} & 98.1 $\pm$ 1.9 (51.6) & 97.3 $\pm$ 0.9 (108.1) & 95.9 $\pm$ 4.7 (20.4) & 83.1 $\pm$ 1.7 (48.9) & 84.1 $\pm$ 2.2 (38.2) & 60.8 $\pm$ 4.2 (14.5) \\
 \addlinespace
\textit{RMDT-$L_{1}LR$} & 91.1 $\pm$ 3.1 (29.4) & 83.4 $\pm$ 4.1 (20.3) & 90.1 $\pm$ 4.2 (21.5) & 63.1 $\pm$ 4.2 (15.0) & 76.6 $\pm$ 2.7 (28.4) & 94.1 $\pm$ 1.4 (67.2) & 75.9 $\pm$ 5.9 (12.9) \\
\textit{RMAR-$L_{1}LR$} & 94.5 $\pm$ 1.7 (55.6) & 96.6 $\pm$ 0.9 \textbf{(107.3)} & 91.2 $\pm$ 5.6 (16.3) & 62.1 $\pm$ 5.5 (11.3) & 81.4 $\pm$ 5.1 (16.0) & 92.4 $\pm$ 2.7 (34.2) & 67.0 $\pm$ 4.5 (14.9) \\
\textit{RM1D-$L_{1}LR$} & 96.2 $\pm$ 1.3 (74.0) & 97.0 $\pm$ 1.4 (69.3) & 97.3 $\pm$ 0.9 (108.1) & 97.4 $\pm$ 2.9 (33.6) & 83.1 $\pm$ 1.5 (55.4) & 83.3 $\pm$ 2.3 (36.2) & 58.6 $\pm$ 3.7 (15.8) \\
 \addlinespace
\textit{RMDT-SVM lin} & 90.6 $\pm$ 3.2 (28.3) & 85.8 $\pm$ 3.2 (26.8) & 90.1 $\pm$ 4.2 (21.5) & 66.4 $\pm$ 5.5 (12.0) & 76.6 $\pm$ 2.7 (28.4) & 94.7 $\pm$ 1.2 (78.9) & 77.5 $\pm$ 4.4 (17.6) \\
\textit{RMAR-SVM lin} & 94.8 $\pm$ 1.7 (55.8) & 96.1 $\pm$ 2.1 (45.8) & 92.5 $\pm$ 5.0 (18.5) & 67.5 $\pm$ 4.4 (15.3) & \textcolor{red}{85.1} $\pm$ 2.9 (29.3) & 90.7 $\pm$ 4.9 (18.5) & 69.0 $\pm$ 1.5 \textbf{(46.0)} \\
\textit{RM1D-SVM lin} & 96.4 $\pm$ 0.9 (107.1) & 98.0 $\pm$ 2.6 (37.7) & \textcolor{red}{98.2} $\pm$ 0.9 \textbf{(109.1)} & \textcolor{red}{97.9} $\pm$ 0.6 \textbf{(163.2)} & 82.7 $\pm$ 2.2 (37.6) & 88.0 $\pm$ 0.8 \textbf{(110.0)} & 69.2 $\pm$ 4.5 (15.4) \\
 \addlinespace
\textit{RMDT-SVM rbf} & 90.6 $\pm$ 3.2 (28.3) & 85.8 $\pm$ 3.2 (26.8) & 90.1 $\pm$ 4.2 (21.5) & 66.5 $\pm$ 5.3 (12.5) & 76.6 $\pm$ 2.7 (28.4) & 94.7 $\pm$ 1.2 (78.9) & \textcolor{red}{78.3} $\pm$ 3.9 (20.1) \\
\textit{RMAR-SVM rbf} & 94.2 $\pm$ 1.8 (52.3) & 96.1 $\pm$ 2.1 (45.8) & 92.9 $\pm$ 5.4 (17.2) & 65.8 $\pm$ 4.9 (13.4) & 82.6 $\pm$ 1.9 (43.5) & 92.3 $\pm$ 2.3 (40.1) & 67.3 $\pm$ 6.0 (11.2) \\
\textit{RM1D-SVM rbf} & 96.2 $\pm$ 1.0 (96.2) & \textcolor{red}{98.7} $\pm$ 1.2 (82.2) & \textcolor{red}{98.2} $\pm$ 0.9 \textbf{(109.1)} & 90.4 $\pm$ 3.2 (28.2) & 80.2 $\pm$ 4.1 (19.6) & \textcolor{red}{95.6} $\pm$ 1.2 (79.7) & 74.2 $\pm$ 3.9 (19.0) \\
 \addlinespace
\textit{RMDT-Clf var.} & \textbf{0.04} & 1.3 & \textbf{0.0} & \textbf{5.8} & \textbf{0.0} & \textbf{0.1} & \textbf{0.9} \\
\textit{RMAR-Clf var.} & 0.2 & \textbf{0.06} & 1.6 & 8.2 & 2.3 & 0.6 & \textbf{0.9} \\
\textit{RM1D-Clf var.} & 0.1 & 0.4 & 0.2 & 8.9 & 1.3 & 23.7 & 39.7 \\
 \addlinespace
\textit{RMDT-Clf* stab.} & 51.7 $\pm$ 35.4 & 49.2 $\pm$ 36.5 & 65.7 $\pm$ 26.4 & 70.7 $\pm$ 21.4 & 57.1 $\pm$ 31.9 & 95.6 $\pm$ 4.8 & 95.5 $\pm$ 4.1 \\
\textit{RMAR-Clf* stab.} & 91.8 $\pm$ 6.1 & 76.4 $\pm$ 17.0 & 83.0 $\pm$ 12.4 & 85.2 $\pm$ 11.0 & 79.5 $\pm$ 15.3 & 91. $\pm$ 7.2  & 86.5 $\pm$ 10.0 \\
\textit{RM1D-Clf* stab.} & 90.6 $\pm$ 6.8 & 83.2 $\pm$ 12.6 & 87.8 $\pm$ 11.2 & 96.7 $\pm$ 2.9 & 85.3 $\pm$ 12.8 & 88.9 $\pm$ 9.8 & 77.9 $\pm$ 16.9 \\
\bottomrule
\addlinespace
\multicolumn{7}{@{}l}{}\\
\end{tabular}}
\caption{\textbf{Performance scores of the models of the 3rd level of comparison.} The table reports the F1 score mean$\pm$sd (mean/std) over the 5 splits. For each dataset, the best model in terms of mean (resp. mean/std) score is highlighted in red (resp. bold). The values listed after the performance scores are the variances of the mean F1 scores over the hybrid (\ie, \textit{RMDR-Clf, RMAR-Clf}) and our local models (\ie, \textit{RM1D-Clf var.}). At the bottom of the table, we compared the stability (mean$\pm$sd) of the best models for each dataset. The stability is defined as the common percentage of selected local features between the models over the 5 splits}
\label{results_F1score_3}
\end{sidewaystable}

\begin{table}
\centering
\small
\resizebox{\textwidth}{!}{\begin{tabular}{@{} l *{9}{c} @{}}
\toprule
\textbf{Benchmark} & \multicolumn{7}{c@{}}{\textbf{Datasets}}\\
\cmidrule(l){2-8}
& \textbf{WDBC} & \textbf{Wine} & \textbf{Iris} & \textbf{Balance scale} & \textbf{Heart disease} & \textbf{Synthetic} & \textbf{Synthetic noisy} \\
\midrule
\textit{RF}, \textit{GBT}, $L_{2}$ \textit{Clf} & 30 & 13 & 4 & 20 & 22 & 6 & 6 \\
\textit{RMDT-$L_{2} Clf$} & 6 & 7 & 8 & 33 & 10 & 19 & 41 \\
\textit{RMAR-$L_{2} Clf$} & 884 & 170 & 51 & 134 & 1029 & 128 & 111 \\
\textit{RM1D-$L_{2} Clf$} & 56 & 36 & 11 & 36 & 25 & 11 & 11 \\
 \addlinespace
$L_{1}LR$ & 20 & 13 & 4 & 20 & 18 & 6 & 6 \\
\textit{RMDT-$L_{1}LR$} & 6 & 7 & 6 & 33 & 9 & 18 & 34 \\
\textit{RMAR-$L_{1}LR$} & 126 & 102 & 21 & 111 & 11 & 69 & 85 \\
\textit{RM1D-$L_{1}LR$} & 18 & 20 & 10 & 36 & 14 & 10 & 11 \\
 \addlinespace
\bottomrule
\addlinespace
\multicolumn{7}{@{}l}{}\\
\end{tabular}}
\caption{\textbf{Complexity of the 18 different tested models for all the datasets.} This table reports the median complexity of the models over the 5 splits, defined as the number of input features (original or derived rules) used within the models. $L_{2}Clf$ corresponds to $L_{2}LR$, \textit{SVM lin} and \textit{SVM rbf} methods.
}
\label{results_complexity}
\end{table}

In this first level of comparison we aimed at comparing our local predictive models (\textit{RM1D-Clf}) with the classifiers (\textit{Clf}) used in our local predictive models (\ie, global predictive models) on the real-life and synthetic datasets. For the predictive performance, we both reported the mean, std and mean/std of the F1 scores over the 5 splits for a more relevant comparison. 

If we just look at the average accuracy (\ie, mean F1 scores), our local models outperformed the global ones for all the datasets except for the balance scale and heart disease datasets for which our local predictive models achieved similar average predictive scores to those of the global models, \ie, 98.2 (97.9) with \textit{SVM lin} (\textit{RM1D-SVM lin}) for balance scale and 83.3 (83.1) with $L_{2}LR$ (\textit{RM1D-$L_2LR$}) for heart disease (see table ~\ref{results_F1score_1}). However, the best mean/std F1 scores were achieved by our local method for all the datasets due to a small variation between the models over the 5 splits (\ie, std F1 scores), highlighting its predictive power and stability.   

It is also noteworthy that the variance of the mean F1 scores between the local predictive models was considerably lower than the variation of the performances between the global models for all the real datasets (see \textit{Clf var.} and \textit{RM1D-Clf var.} in table~\ref{results_F1score_1}). The opposite result was observed for the synthetic datasets due to the \textit{RM1D-SVM rbf} local model being able to efficiently increase the average classification accuracies of datasets built based on decision rules.
These results suggest that on real data, our rule-mining step providing local features yields more predictive power and more robust input to the learning process of the standard classifiers. Conversely, on synthetic data based on decision rules, the choice of the classifier applied to the local features have a great impact on predictive accuracy.  

Our local models had more final features than the global ones due to the exhaustive local exploration strategy of the rule-mining algorithm. Moreover, we might also conclude that our rule selection process is efficient since the final number of local features within the local models remains of the same magnitude as those within the global models,  whatever the penalization used (see table~\ref{results_complexity}). Moreover, our local predictive models provide more precise, interpretable and meaningful information through the use of rules.

In conclusion, the first step of comparison demonstrated the added-value of our rule-mining algorithm prior to the classification step in terms of performance, interpretability, complexity and stability.

\subsection{2nd level of comparison: "Classification Rule Learning Methods \textit{vs} our Local Predictive Models"}

Secondly, we aimed at comparing our local predictive method (\textit{RM1D-Clf}) with well-known classification rule learning methods: \textit{RF} and \textit{GBT}.

If we only consider at the average accuracy (\ie, mean F1 scores), our local models outperformed the two classification rule learning models except for the wine and synthetic noisy datasets for which our local predictive models achieved almost identical average predictive scores to those of the global models, \ie, 99.0 (98.7) with \textit{RF} (\textit{RM1D-Clf*}, \ie, \textit{RM1D-SVM rbf}) for wine and 75.7 (74.2) with \textit{GBT} (\textit{RM1D-Clf*}, \ie, \textit{RM1D-SVM rbf}) for synthetic noisy (see table ~\ref{results_F1score_2}). However, the best mean/std F1 scores were achieved by our local method for all the datasets due to a small variation between the models over the 5 splits (\ie, std F1 scores), highlighting again its predictive power and stability. We also observed that both \textit{RF} and \textit{GBT} completely failed to predict the outcome of the balance scale dataset probably because of the linearity of the final decision. Our local model (\ie, \textit{RM1D-SVM lin)} outperforms them thanks to its combined strategy.

Again, our local models generally had more final features than the hybrid local ones but remains of the same magnitude (see table~\ref{results_complexity}). Moreover, if we consider the number of trees (at least 100) and consequently the number of "rules" used by \textit{RF} and \textit{GBT} for the outcome prediction, our model becomes much less complex. They are equally much more difficult to interpret than our model.

To conclude this second level of comparison, our local models allowed the extraction of one single set of individual (one dimensional) rules which are equally or more discriminative than the best set of multiple and complex rules extracted by \textit{RF} or \textit{GBT}.

\subsection{3rd level of comparison: "Hybrid \textit{vs} our Local Predictive Models"}
Finally, we aimed at comparing, in terms of predictive performance, complexity and stability, our local predictive method (\textit{RM1D-Clf}) with two other hybrid local predictive methods using rules from decision tree (\textit{RMDT-Clf}) or association rules (\textit{RMAR-Clf}) techniques as input data to the classifiers.

Except for heart disease and synthetic noisy datasets, the best models regarding the mean F1 score were achieved with our local predictive method. For these two datasets, our local models still got average performances of the same order than the hybrid ones based on \textit{RMAR} (see table ~\ref{results_F1score_3}). 

For heart disease, the hybrid method based on \textit{RMAR} always yielded the best models in terms of average classification accuracy except when combined to \textit{$L_{1}LR$} classifier where our local method had the best performance. We can assume that the \textit{RMAR} strategy based on frequent item set discovery suit well the medical context and goal of the dataset, and also the definition of the original features and their values. We also noted that \textit{RMAR} combined with non-sparse classifiers led to models 40 to 100 times more complex than the two other local methods. Besides, our local method outperformed the two hybrid methods when combined to $L_{1}LR$ classifier for all the real datasets. We can assume that the $L_{1}$ penalization works better with our local features on real datasets highlighting their discriminative power.

Unsurprisingly, for the two synthetic datasets \textit{RMDT-Clf} largely outperformed whatever the classifier used except when combined to \textit{SVM rbf} with which our local method significantly improved its predictive capacity. Indeed, \textit{RM1D-SVM rbf} outperformed the hybrid on the non-noisy Synthetic dataset (see table ~\ref{results_F1score_3}) and increased closer to the best mean and mean/std F1 scores on the synthetic noisy dataset. These results  on synthetic datasets (and the previous ones in tables ~\ref{results_F1score_1} and ~\ref{results_F1score_2}) might suggest that a non-linear combination of our 1D local features is useful to predict outcome based on rules. 

It appeared that \textit{RM1D-Clf} is the only strategy able to classify correctly for the balance scale dataset for which we observed also the strongest variances of the mean F1 scores for all the strategies through all the datasets (see table~\ref{results_F1score_3}). For all the other datasets, we got small variances of the mean F1 scores for the hybrid models (see \textit{RMDT-Clf var.} and \textit{RMAR-Clf var.} in table~\ref{results_F1score_3}). The same results were observed for \textit{RM1D-Clf var.} except for the two synthetic datasets where the use of linear and non-linear \textit{SVM} classifiers significantly increase the predictive power. 

Our best local predictive models were better and more stable (\ie, same features across the 5 splits) than the best hybrid models for all the real datasets except for WDBC dataset for which \textit{RMAR-} and \textit{RM1D-Clf*} were equally stable (see \textit{RMDT-Clf* stab.}, \textit{RMAR-Clf* stab.} and \textit{RM1D-Clf* stab.} in table~\ref{results_F1score_3}). In accordance with its performance score, the stability of our best local predictive model on the synthetic noisy dataset is much less stable than that of the hybrid ones.   

To conclude this third level of comparison, we show by assessing different strategies of rules' extraction (decision tree, association rules and our rule-mining algorithm) that our rule-mining algorithm is a good strategy in general. Indeed, the classifiers had in most cases superior performances when they were applied to the local features extracted by our rule-mining algorithm. Moreover, our local models were much less complex than the hybrid-based association rules models and much more stable than the hybrid-based decision tree models. Moreover, our local models provided a synthetic and easily interpretable information under the form of a predictive combination of weighted 1D rules instead of a combination of weighted multi-dimensional rules.

\subsection{From rules to insights}
For each dataset, we reviewed the insights yielded by the rules generated by our local predictive models. 

\begin{figure}[!htb]
\centering
\begin{subfigure}{1.0\textwidth}
\centering
\includegraphics[width=0.8\textwidth]{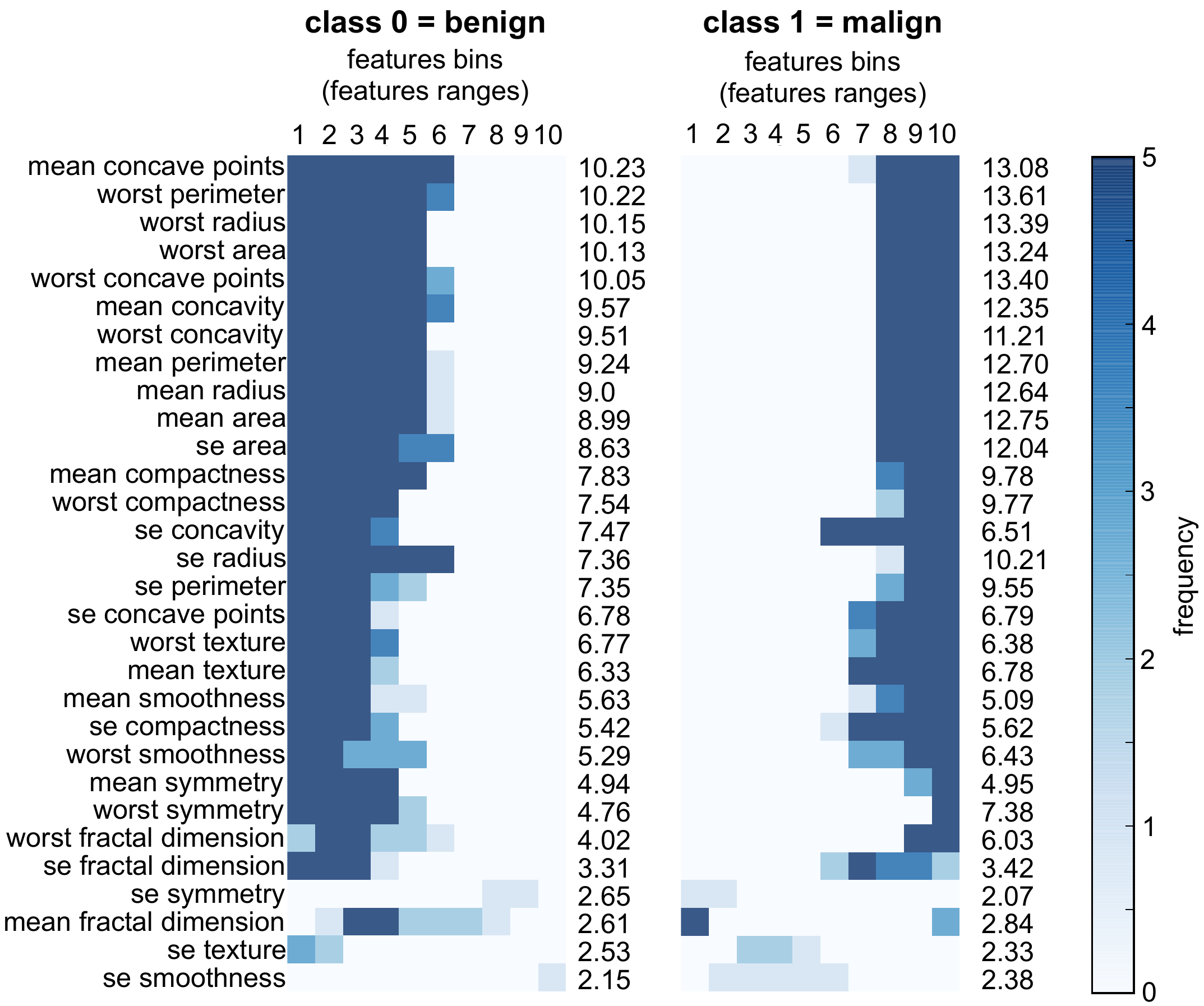}
\caption{WDBC dataset} \label{rules_wdbc}
\end{subfigure}
\begin{subfigure}{1.0\textwidth}
\centering
\includegraphics[width=0.8\textwidth]{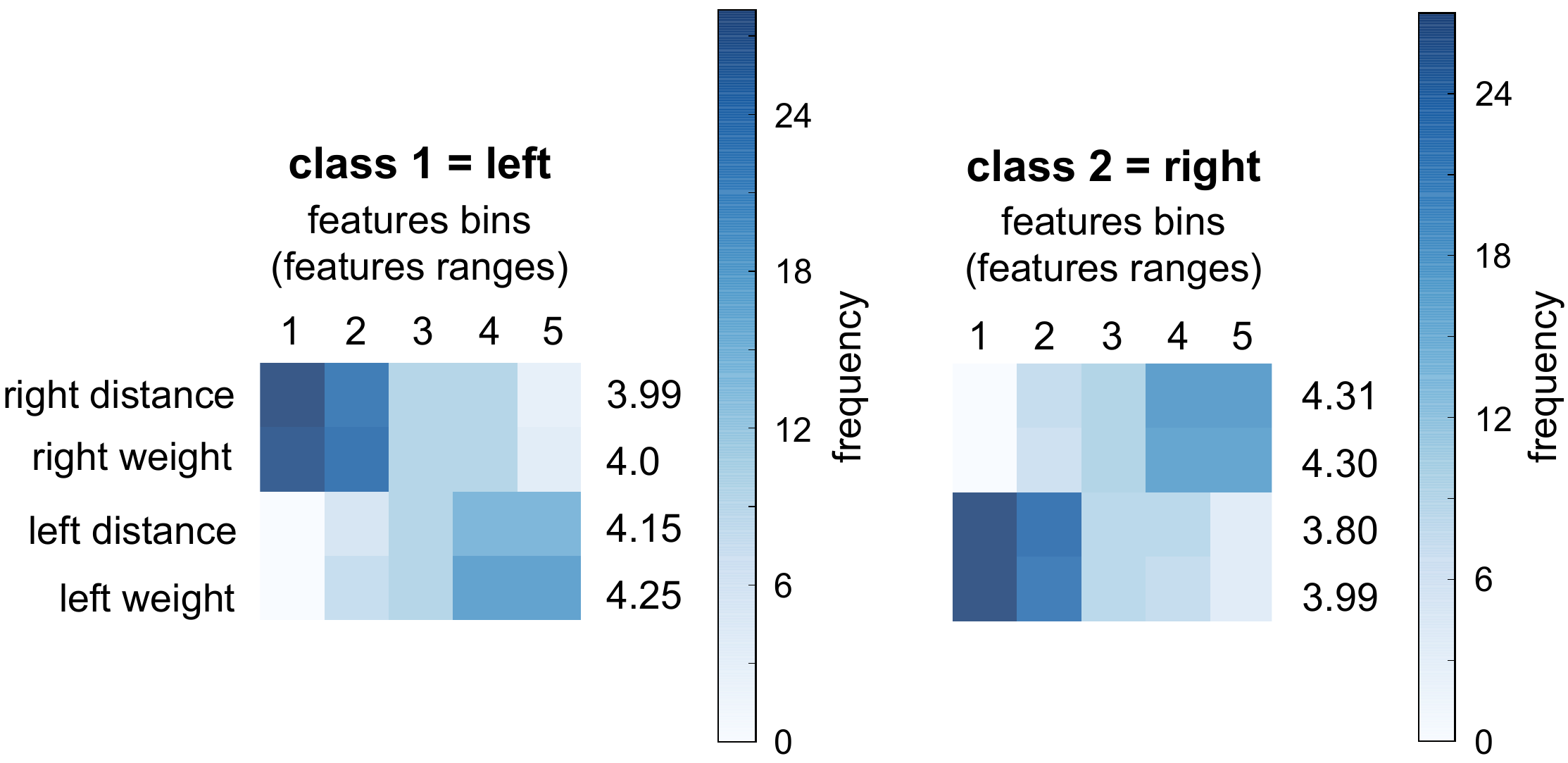}
\caption{Balance scale dataset} \label{rules_balance_scale}
\end{subfigure}
\caption{\textbf{Representation of the 1D rules for the a) WDBC and b) balance scale datasets.}
Each horizontal segment corresponds to a 1D rule characterized by its feature condition: the feature’s name and the set of covered values or bins. The color scale reflects the frequency of the features’ values covered by the rules over the 5 splits. The more robust the rule, the darker it will be. On the left figures, the rules corresponding to the a) benign class and b) left class and on the right figures, the rules corresponding to the a) malign class and b) right class. For each feature and each class, the average z-score of the related rules is given on the right side of each figure.}
\end{figure}

For the \textbf{WDBC dataset}, at least one significant rule was selected for each one of the original features and for each class (see figure~\ref{rules_wdbc}). Some rules were always selected (see the rule frequency) and provided high discriminative information regarding their z-score. Moreover, we noted that the original feature listed in decreasing order according to their average z-score was approximately the same from one class to another. In addition, the feature condition of rules related to the same original feature but to different classes are disjoint. Therefore the feature space is covered by disjoint rules, \ie, distinct sub-regions, over classes which provides a simple interpretation without ambiguities. These disjoint rules obtained with our local models indicate that the benign class was mainly characterized by low values of the features (\ie, low feature condition) and that conversely the malign class was mainly characterized by high values of the features. These results are in accordance with the literature [\cite{tan2003evolutionary}].

For the \textbf{balance scale dataset}, the correct way to find the class is the greater of (\textit{left distance} * \textit{left weight}) and (\textit{right distance} * \textit{right weight}). If they are equal, it is balanced. Our local models find a set of rules (see figure~\ref{rules_balance_scale}) on all the original features but only on two of the three classes, corresponding to left and right classes. The average z-score were approximately the same whatever the feature considered demonstrating that they all contribute to the prediction. There is understandably an overlap of all the rules (over the 5 splits) of the left and right classes having centered feature condition. But these rules are less frequently selected over the 5 splits than those located at the left and right ends having more disjoint feature condition between the two classes. Indeed, samples with higher values of \textit{left distance} and \textit{left weight} and lower value of \textit{right distance} and \textit{right weights} are more likely selected as representative of the left class and conversely for the right class. No rule was found for the balanced class for which samples must have \textit{right distance} $=$ \textit{left distance} and \textit{right weight} $=$ \textit{left weight}. This is because samples of this class (only 7.84 $\%$ of the total number of samples) can be spread across all values making difficult the discovery of a subgroup of samples with rules. However, the linear decision functions learned by the linear classifiers based on the rules of the left and right classes were efficient to well classify the samples into the three classes.

\begin{figure}[!htb]
\centering
\begin{subfigure}{1.0\textwidth}
\centering
\includegraphics[width=0.8\textwidth]{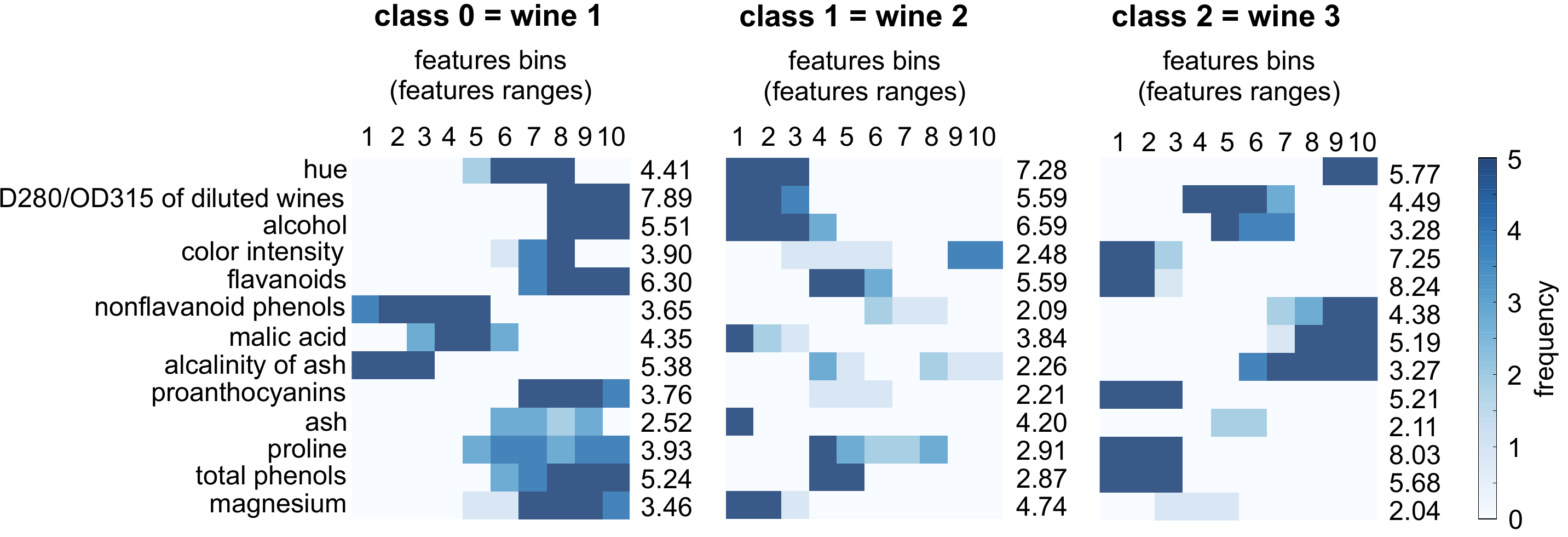}
\caption{Wine dataset} \label{rules_wine}
\end{subfigure}
\begin{subfigure}{1.0\textwidth}
\centering
\includegraphics[width=0.8\textwidth]{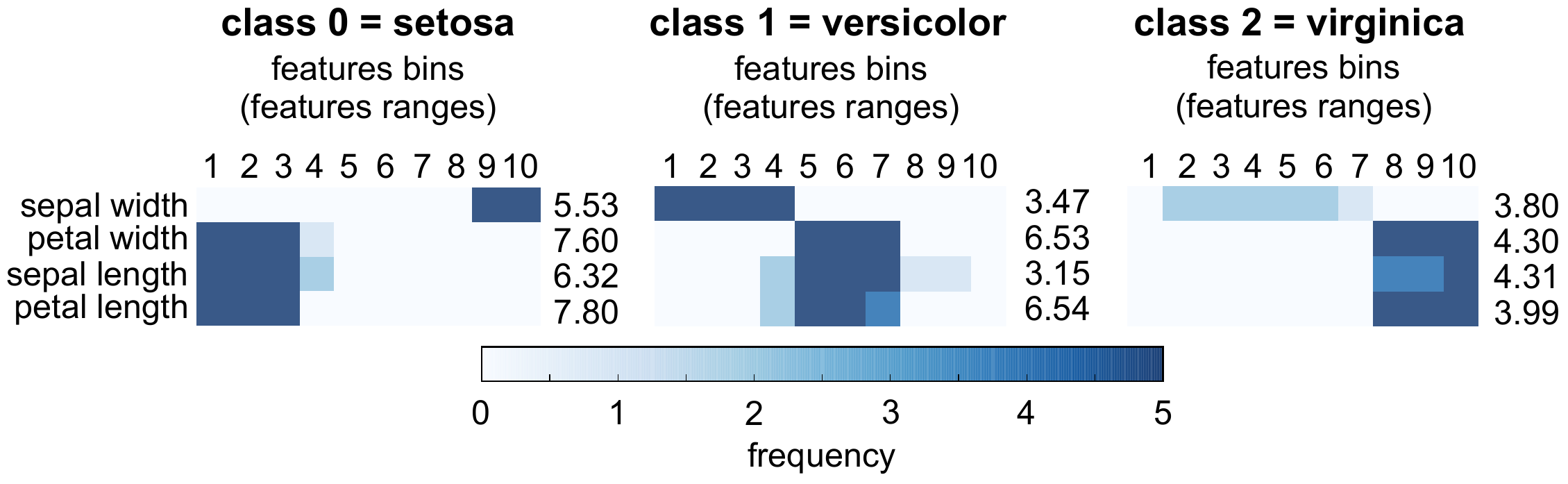}
\caption{Iris dataset} \label{rules_iris}
\end{subfigure}
\caption{\textbf{Representation of the 1D rules for the a) wine and b) iris datasets.}
Each horizontal segment corresponds to a 1D rule characterized by its feature condition: the feature’s name and the set of covered values or bins. The color scale reflects the frequency of the features’ values covered by the rules over the 5 splits. The more robust the rule, the darker it will be. On the left figures, the rules corresponding to the a) wine 1 class and the b) iris-setosa class, on the middle figures, the rules corresponding to the a) wine 2 class and the b) iris-versicolor class, on the right figures, the rules corresponding to the a) wine 3 class and the b) iris-virginica class. For each feature and each class, the average z-score of the related rules is given on the right side of each figure.}
\end{figure}

For the \textbf{wine dataset}, as for WDBC dataset, at least one significant rule was selected for each one of the original features and for each class (see figure~\ref{rules_wine}). For each feature, the most frequent rules (\ie, selected at each one of the 5 splits) are always disjoint between the three classes. The original feature of the rule with the highest average z-scores are different for the three classes and therefore the type of wines: \textit{D280/0D315 of diluted wine} for Wine 1, \textit{hue} for Wine 2 and \textit{color intensity} for Wine 3. This is a well posed problem with specific class structures that our models managed to extract. We clearly visualize these class structures through the mainly disjoint rules (z-score $>$ 5). Indeed, these disjoint rules indicate that 1) wine 1 is mostly characterized by high values of \textit{D280/0D315 of diluted wine}, \textit{alcohol}, \textit{flavanoids} and \textit{total phenols}, low values of \textit{acalinity of ash}; 2) wine 2 is mostly characterized by low values of \textit{hue}, \textit{D280/0D315 of diluted wine}, \textit{alcohol} and \textit{ash},  and middle values of \textit{flavanoids}; 3) wine 3 is mostly characterized by low values of \textit{color intensity}, \textit{flavanoids}, \textit{proanthocyaninc}, \textit{proline} and \textit{total phenols}, and high values of \textit{hue} and \textit{malic acid}. 

The \textbf{iris dataset} consists in four features describing three types of iris.
Based on the values observed in the datasets, the following 1 to 3 dimensional rules might be learned from this dataset [\cite{witten2016data}]:
\begin{align}
  class=
  \begin{cases}
    \text{iris-setosa}  & \text{if }  {\textit{petal length} < 2.5} \\\\
    \text{iris-versicolor} & \text{if }  
    \begin{array}{l}
    	{\textit{sepal width} < 2.10}\\
        {\textit{sepal width} < 2.45, \textit{ petal length} < 4.55}\\
        {\textit{sepal width} < 2.95, \textit{ petal width} < 1.35}\\
        {\textit{petal length} \geq 2.45, \textit{ petal length} < 4.45}\\
        {\textit{sepal length} \geq 5.85, \textit{ petal length} < 4.75}\\
        {\textit{sepal width} < 2.55, \textit{ petal length} < 4.95, \textit{ petal width} < 1.55}\\
        {\textit{petal length} \geq 2.45, \textit{ petal length} < 4.95, \textit{ petal width} < 1.55}\\
        {\textit{sepal length} \geq 6.55, \textit{ petal length} < 5.05}\\
        {\textit{sepal width} < 2.75, \textit{ petal width} < 1.65, \textit{ sepal length} < 6.05}\\
        {\textit{sepal length} \geq 5.85, \textit{ sepal length} < 5.95, \textit{ petal length} < 4.85}\\\\
	\end{array}\\
   \text{iris-virginica} & \text{if }
    \begin{array}{l}
    	{\textit{petal length} \geq 5.15}\\
        {\textit{petal width} \geq 1.85}\\
        {\textit{petal width} \geq 1.75, \textit{ sepal width} < 1.55}\\\\
	\end{array}\\
  \end{cases}
  \label{iris_dataset}
\end{align}

These rules are very cumbersome. Our local rule-mining algorithm learned more compact rules having a strong predictive power (see~\ref{rules_iris}). At least one significant rule was selected for each one of the features and for each class. Most of the rules are selected for all the train-test splits and disjoint between the three classes. We observed an overlap between iris-versicolor and iris-virginica classes for the rules related to the features \textit{sepal- width} and \textit{length}. These overlapping rules have lower z-score and frequency than the disjoint rules. This is in accordance with the literature results. Indeed, the iris dataset is known to contain two linearly separable clusters: one containing the iris-setosa class and another containing both iris-virginica and iris-versicolor class and only a small fraction of iris-virginica is mixed with iris-versicolor [\cite{fisher1936use,gorban2007topological,gorban2010principal}]. Despite this overlap within our 1D rules, our models were able to discriminate the three types of iris. This is probably due to our rule-mining strategy combined with a classifier that builds an efficient decision function based on the linear or non-linear combination of the 1D rules.

\begin{figure}[!htb]
\centering
\includegraphics[width=0.6\textwidth]{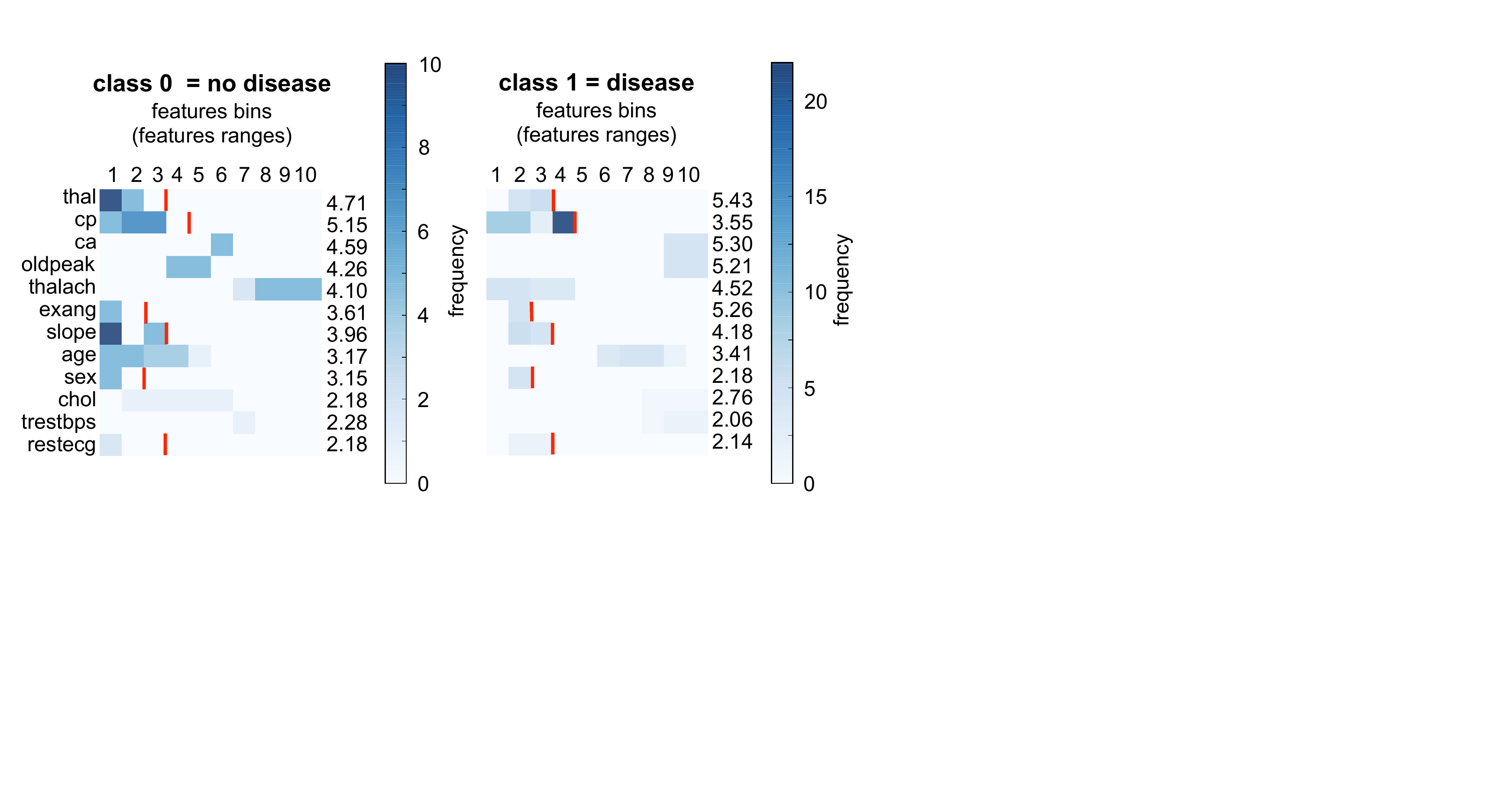}
\caption{\textbf{Representation of the 1D rules for the heart disease dataset.}
Each horizontal segment corresponds to a 1D rule characterized by its feature condition: the feature’s name and the set of covered values or bins. The color scale reflects the frequency of the features’ values covered by the rules over the 5 splits. As discrete features do not have the same number of bins as continuous features, we indicated with a red line the corresponding upper bin of these features. The more robust the rule, the darker it will be. On the left figure, the rules corresponding to the no heart disease class and on the right figure the rules corresponding to the disease class. For each feature and each class, the average z-score of the related rules is given on the right side of each figure.}
\label{rules_heart_disease}
\end{figure}

The \textbf{heart disease dataset} has 13 mixed features that are potential risk factor for having or not heart disease. At least one rule was selected for each one of the original features for at least one of the two classes except for the feature \textit{fbs} were no rule exists (see figure~\ref{rules_heart_disease}). Noted that for disease class, the rule on the bin 4 of the feature \textit{cp} was always selected over the 5 splits and represents for almost the third of the rules selected for this class. Besides, the mainly disjoint and significant rules which are mildly frequent were related to the \textit{exang}, \textit{thalach}, \textit{age}, \textit{sex}, \textit{oldpeak} and \textit{ca} features. Our results largely confirm the strong contribution of these features and corresponding rules to the diagnosis of the disease as reviewed in several studies. For instance, the decreasing order of our z-scores is very close to that of the intrinsic discrepancies between the disease and no-disease distributions of the features computed in \url{http://lucdemortier.github.io/projects/3_mcnulty}. Some of the most relevant rules have been also reported in [\cite{el2015feature}]. 

\begin{figure}[!htb]
\centering
\begin{subfigure}{1.0\textwidth}
\centering
\includegraphics[width=0.8\textwidth]{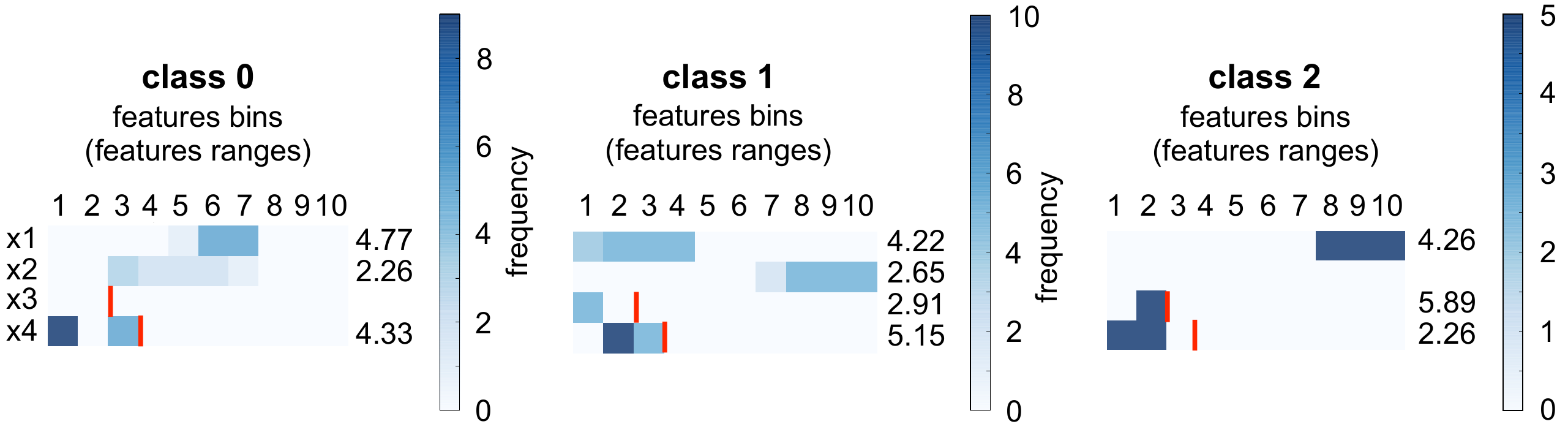}
\caption{Synthetic dataset} \label{rules_synthetic}
\end{subfigure}
\begin{subfigure}{1.0\textwidth}
\centering
\includegraphics[width=0.8\textwidth]{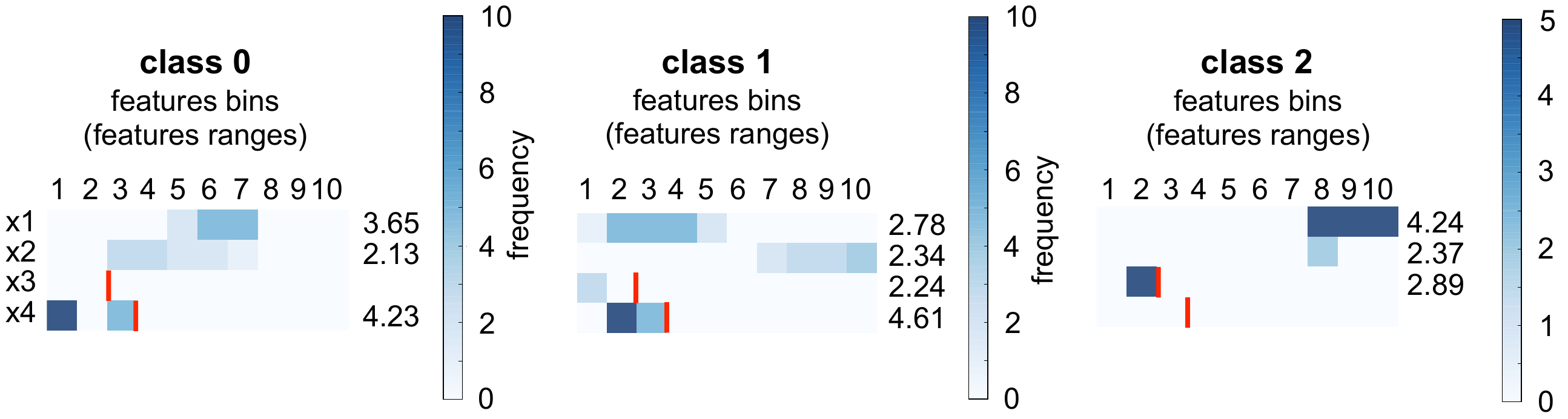}
\caption{Synthetic noisy dataset} \label{rules_synthetic_noisy}
\end{subfigure}
\begin{subfigure}{1.0\textwidth}
\centering
\includegraphics[width=0.8\textwidth]{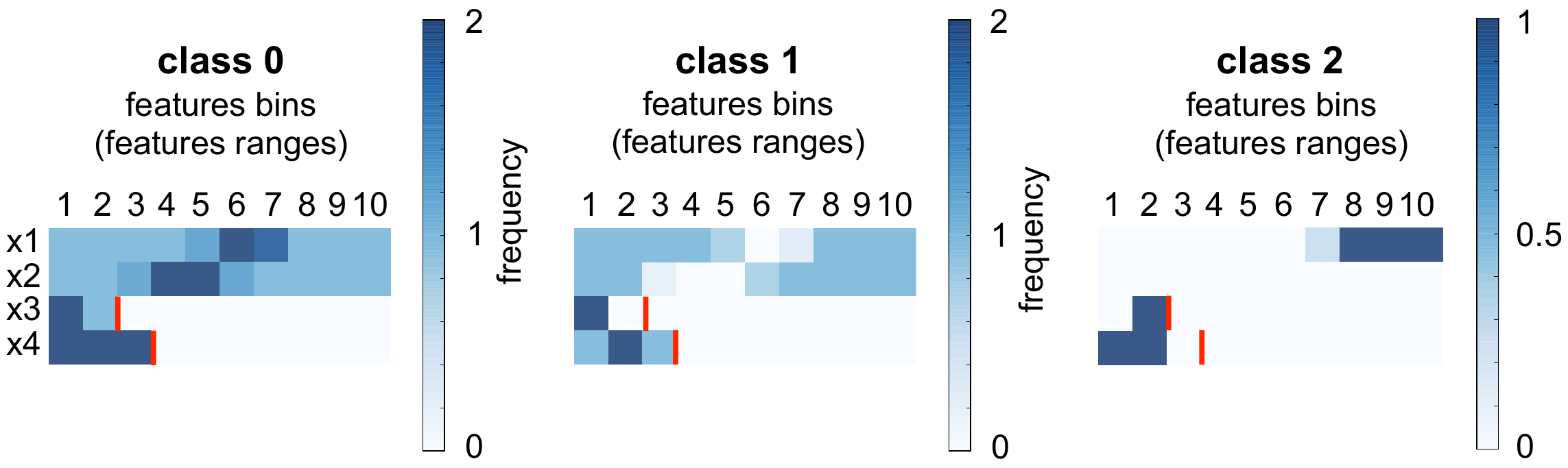}
\caption{Decomposition in 1D rules of the decision tree rules used to construct the synthetic datasets} \label{rules_synthetic_true}
\end{subfigure}
\caption{\textbf{Representation of the 1D rules for the two synthetic datasets.}
Each horizontal segment corresponds to a 1D rule characterized by its feature condition: the feature’s name and the set of covered values or bins. The color scale reflects the frequency of the features’ values covered by the rules and over the 5 splits for figure a and b. As discrete features do not have the same number of bins as continuous features, we indicated with a red line the corresponding upper bin of these features. The more robust the rule, the darker it will be. On the left figure, the rules corresponding to the class 0, on the middle figure, the rules corresponding to the class 1 and, on the left figure, the rules corresponding to the class 2. For figure a and b, for each feature and each class, the average z-score of the related rules is given on the right side of each figure.}
\end{figure}

For the \textbf{synthetic dataset} (see figure~\ref{rules_synthetic}), class 2 is perfectly characterized by three consistent 1D rules which, when combined, give the original true rule. For class 1, our selected 1D rules recover the original first and third true rules. For class 0, despite missing feature conditions within the selected 1D rules, the first, second and fourth original true rules were recovered because the missing feature conditions within these three true rules ($x_{3}$/$x_{4}$ for the $1^{st}$ and $2^{nd}$/$4^{th}$ rule) are finally useless to discriminate class 0 from the others. The fifth true rule seems to be partially recovered by two 1D rules: $x_{4} = \text{blue}$ and $x_{1} \in [0.503, 0.715]$. But, in fact this true rule might be totally recovered thanks to the learned decision boundaries based on the disjoint rules of the $x_{1}$ feature between the three classes. Similar arguments might hold for the third and second true rules of the classes 0 and 1 respectively. Indeed, if $x_{4} = \text{blue}$ or white and $x_{1} \geq 0.7$, then samples could fall into one of the three classes according to the value of $x_{3}$ and $x_{2}$. The single 1D rule of class 2 effectively discriminates samples of class 2 from the others according to $x_{3}$. Unfortunately, as can be seen in figure~\ref{rules_synthetic_true}, 1D rules derived from the split of the original multi-dimensional rules lead to concurrent 1D rules for the feature $x_{2}$. Our rule mining strategy chose to select disjoint 1D rules on $x_{2}$ for the classes 0 and 1 being able to discriminate the first, second and a part of the third true rules of class 0 and the first true rule of class 1. Therefore, no rule allows to detect samples verifying the second true rules of class 1.

For the \textbf{synthetic noisy dataset}, the selected 1D rules of the classes 0 and 1 were the same as those of the non noisy synthetic dataset but they have lower average z-score reflecting a lower signal-to-noise ratio (see figure~\ref{rules_synthetic_noisy}). By contrast, the rules of the class 2 presents some differences: no rule selected on the feature $x_{4}$ and one irrelevant rule was sometimes selected on the feature $x_{2}$. However, the two remaining consistent 1D rules on the features $x_{1}$ and $x_{3}$ are enough to reach a good classification of this class. These differences observed in the 1D rules selected for the non noisy and noisy synthetic dataset probably explains the decrease in the classification performances. We recall that \textit{RMDT-Clf} models were naturally more performant on the two synthetic datasets and also more robust to the addition of noise. However, in real life, it's rare to face data that would match perfectly with a decision tree.

\section{Conclusion}

The principle of a rule-mining approach is to explore subparts of the feature space and to generate rules (\ie, group of features with their sets of values) representative of subgroups of samples having a high density for the outcome of interest. Our rule-mining algorithm has the particularity of being able to explore all possible combinations of subspaces of features without any prior assumptions. Then, the rule selection step provides a set of highly concentrate rules that are in most cases disjoint between the classes facilitating further interpretation without ambiguities. Our benchmark study proved that our local predictive strategy is a good competitor to both resolve classification problems on real-world datasets and provide meaningful explanation. Indeed, for the various real datasets of this study, our challenging method either outperformed the other strategies of the benchmark or achieved similar accuracies. It also achieved competitive results relative to those reported in the literature. Results derived from the first level of comparison demonstrated the benefit of combining both our rule-mining algorithm with a classifier in terms of performance, complexity and stability. Indeed, the standard classifiers seem to learn more discriminative decision boundaries based on the single set of individual rules (\ie, local features) rather than based on the original features. Therefore, our method proposes an original and efficient strategy of projection of the original features to a meaningful space where the final decision function is highly discriminative, and directly readable for linear classifiers. The second and third levels of comparison confirmed the benefits of using single set of individual rules extracted with our rule-mining algorithm rather than the best set of multiple and complex rules extracted by two standard classification rule learning methods (\ie, \textit{RF} or \textit{GBT}) on the one hand, and on the other hand, the use of multi-dimensional rules extracted by two rule-mining methods (\ie, \textit{RMDT} and \textit{RMAR}). With our rule-mining algorithm, we globally observed more predictive, interpreting, stable and simple results. 

Some questions and improvements still remain open regarding the implementation options of some parts of our rule-mining algorithm. Firstly, the minimization part of the rule-mining algorithm might be revisited to remove many redundancies within the final set of rules especially when the number of dimension increases or for discrete features. Revisiting this part of the algorithm should allow us to take into account rules with more than 2 dimensions, which could be useful in the case of decision tree-based real datasets (cf results on synthetic datasets). However, an increase in complexity is hugely time-consuming, leads to a higher risk of false discoveries, and could be more difficult to interpret. Secondly, we would like to investigate other types of discretization process like supervised discretization [\cite{dougherty1995supervised}] which can have an impact on the final performances. In a first instance, we chose a without a priori and unsupervised discretization process (\ie, 10 bins quantization) for continuous features but a more intuitive or guided strategy could improve the predictive capacity of our strategy. Thirdly, we observed that for some datasets, the use of non-linear interactions of our 1D rules is more efficient than the use of linear interactions. Finally, the interest of the exhaustive exploration of our rule-mining algorithm should be weighted against the drawback of time-consuming computing requiring either the use of powerful computer or prior feature selection.

% Acknowledgements should go at the end, before appendices and references
\acks{Special thanks to Tristan Sylvain for useful comments and proofreading.}

% Manual newpage inserted to improve layout of sample file - not
% needed in general before appendices/bibliography.

\newpage

\appendix
\section*{Appendix A}

\begin{table}[H]
\centering
\small
\begin{adjustbox}{max width=\textwidth}
\begin{tabular}{*{6}{c}}
\hline
\textbf{Bins} & \textbf{mean concave points} & \textbf{worst perimeter} & \textbf{worst radius} & \textbf{worst area} & \textbf{worst concave points} \\
\hline
1 & $[0.0, 0.011[$ & $[50.41, 72.077[$ & $[7.93, 11.22[$ & $[185.2, 384.288[$ & $[0.0, 0.038[$ \\
2 & $[0.011, 0.018[$ & $[72.077, 81.395[$ & $[11.22, 12.487[$ & $[384.288, 475.872[$ & $[0.038, 0.058[$ \\
3 & $[0.018, 0.023[$ & $[81.395, 86.287[$ & $[12.487, 13.312[$ & $[475.872, 544.116[$ & $[0.058, 0.072[$ \\
4 & $[0.023, 0.028[$ & $[86.287, 91.296[$ & $[13.312, 14.003[$ & $[544.116, 599.58[$ & $[0.072, 0.084[$ \\
5 & $[0.028, 0.034[$ & $[91.296, 97.66[$ & $[14.003, 14.97[$ & $[599.58, 686.5[$ & $[0.084, 0.1[$ \\
6 & $[0.034, 0.048[$ & $[97.66, 105.776[$ & $[14.97, 16.008[$ & $[686.5, 781.732[$ & $[0.1, 0.122[$ \\
7 & $[0.048, 0.065[$ & $[105.776, 115.9[$ & $[16.008, 17.388[$ & $[781.732, 927.704[$ & $[0.122, 0.151[$ \\
8 & $[0.065, 0.084[$ & $[115.9, 133.5[$ & $[17.388, 20.316[$ & $[927.704, 1269.0[$ & $[0.151, 0.178[$ \\
9 & $[0.084, 0.101[$ & $[133.5, 158.076[$ & $[20.316, 23.687[$ & $[1269.0, 1677.8[$ & $[0.178, 0.209[$ \\
10 & $[0.101, 0.201[$ & $[158.076, 251.2[$ & $[23.687, 36.04[$ & $[1677.8, 4254.0[$ & $[0.209, 0.291[$ \\
\hline
\hline
\textbf{Bins} & \textbf{mean concavity} & \textbf{worst concavity} & \textbf{mean perimeter} & \textbf{mean radius} & \textbf{mean area} \\
1 & $[0.0, 0.014[$ & $[0.0, 0.045[$ & $[43.79, 65.782[$ & $[6.981, 10.26[$ & $[143.5, 321.6[$ \\
2 & $[0.014, 0.025[$ & $[0.045, 0.092[$ & $[65.782, 73.285[$ & $[10.26, 11.362[$ & $[321.6, 396.524[$ \\
3 & $[0.025, 0.034[$ & $[0.092, 0.137[$ & $[73.285, 77.336[$ & $[11.362, 12.005[$ & $[396.524, 443.604[$ \\
4 & $[0.034, 0.045[$ & $[0.137, 0.177[$ & $[77.336, 81.927[$ & $[12.005, 12.722[$ & $[443.604, 496.416[$ \\
5 & $[0.045, 0.062[$ & $[0.177, 0.227[$ & $[81.927, 86.24[$ & $[12.722, 13.37[$ & $[496.416, 551.1[$ \\
6 & $[0.062, 0.086[$ & $[0.227, 0.287[$ & $[86.24, 91.426[$ & $[13.37, 14.059[$ & $[551.1, 609.836[$ \\ 
7 & $[0.086, 0.112[$ & $[0.287, 0.35[$ & $[91.426, 98.212[$ & $[14.059, 15.058[$ & $[609.836, 701.404[$ \\
8 & $[0.112, 0.15[$ & $[0.35, 0.42[$ & $[98.212, 111.752[$ & $[15.058, 17.075[$ & $[701.404, 917.184[$ \\
9 & $[0.15, 0.203[$ & $[0.42, 0.574[$ & $[111.752, 129.1[$ & $[17.075, 19.53[$ & $[917.184, 1185.56[$ \\
10 & $[0.203, 0.427[$ & $[0.574, 1.252[$ & $[129.1, 188.5[$ & $[19.53, 28.11[$ & $[1185.56, 2501.0[$ \\
\hline
\hline
\textbf{Bins} & \textbf{se area} & \textbf{mean compactness} & \textbf{worst compactness} & \textbf{se concavity} & \textbf{se radius} \\
1 & $[6.802, 13.136[$ & $[0.019, 0.05[$ & $[0.027, 0.094[$ & $[0.0, 0.008[$ & $[0.112, 0.183[$ \\
2 & $[13.136, 16.64[$ & $[0.05, 0.06[$ & $[0.094, 0.126[$ & $[0.008, 0.013[$ & $[0.183, 0.219[$ \\ 
3 & $[16.64, 19.021[$ & $[0.06, 0.07[$ & $[0.126, 0.161[$ & $[0.013, 0.017[$ & $[0.219, 0.246[$ \\
4 & $[19.021, 21.061[$ & $[0.07, 0.08[$ & $[0.161, 0.185[$ & $[0.017, 0.02[$ & $[0.246, 0.28[$ \\
5 & $[21.061, 24.53[$ & $[0.08, 0.093[$ & $[0.185, 0.212[$ & $[0.02, 0.026[$ & $[0.28, 0.324[$ \\
6 & $[24.53, 29.333[$ & $[0.093, 0.109[$ & $[0.212, 0.251[$ & $[0.026, 0.031[$ & $[0.324, 0.37[$ \\ 
7 & $[29.333, 38.466[$ & $[0.109, 0.122[$ & $[0.251, 0.304[$ & $[0.031, 0.037[$ & $[0.37, 0.431[$ \\
8 & $[38.466, 54.131[$ & $[0.122, 0.142[$ & $[0.304, 0.368[$ & $[0.037, 0.046[$ & $[0.431, 0.546[$ \\
9 & $[54.131, 92.212[$ & $[0.142, 0.176[$ & $[0.368, 0.448[$ & $[0.046, 0.059[$ & $[0.546, 0.752[$ \\
10 & $[92.212, 542.2[$ & $[0.176, 0.345[$ & $[0.448, 1.058[$ & $[0.059, 0.396[$ & $[0.752, 2.873[$ \\
\hline
\hline
\textbf{Bins} & \textbf{se perimeter} & \textbf{se concave points} & \textbf{worst texture} & \textbf{mean texture} & \textbf{mean smoothness} \\
1 & $[0.757, 1.278[$ & $[0.0, 0.005[$ & $[12.02, 17.776[$ & $[9.71, 14.073[$ & $[0.053, 0.08[$ \\  
2 & $[1.278, 1.516[$ & $[0.005, 0.007[$ & $[17.776, 20.154[$ & $[14.073, 15.652[$ & $[0.08, 0.084[$ \\
3 & $[1.516, 1.74[$ & $[0.007, 0.008[$ & $[20.154, 21.983[$ & $[15.652, 16.84[$ & $[0.084, 0.088[$ \\
4 & $[1.74, 2.039[$ & $[0.008, 0.01[$ & $[21.983, 23.582[$ & $[16.84, 17.911[$ & $[0.088, 0.092[$ \\
5 & $[2.039, 2.287[$ & $[0.01, 0.011[$ & $[23.582, 25.41[$ & $[17.911, 18.84[$ & $[0.092, 0.096[$ \\
6 & $[2.287, 2.591[$ & $[0.011, 0.012[$ & $[25.41, 26.746[$ & $[18.84, 19.969[$ & $[0.096, 0.099[$ \\
7 & $[2.591, 3.053[$ & $[0.012, 0.014[$ & $[26.746, 28.46[$ & $[19.969, 21.277[$ & $[0.099, 0.103[$ \\ 
8 & $[3.053, 3.769[$ & $[0.014, 0.016[$ & $[28.46, 30.915[$ & $[21.277, 22.44[$ & $[0.103, 0.107[$ \\
9 & $[3.769, 5.136[$ & $[0.016, 0.019[$ & $[30.915, 33.708[$ & $[22.44, 24.997[$ & $[0.107, 0.115[$ \\
10 & $[5.136, 21.98[$ & $[0.019, 0.053[$ & $[33.708, 49.54[$ & $[24.997, 39.28[$ & $[0.115, 0.163[$ \\
\hline
\hline
\textbf{Bins} & \textbf{se compactness} & \textbf{worst smoothness} & \textbf{mean symmetry} & \textbf{worst symmetry} & \textbf{worst fractal dimension} \\
1 & $[0.002, 0.009[$ & $[0.071, 0.103[$ & $[0.106, 0.15[$ & $[0.157, 0.226[$ & $[0.055, 0.066[$ \\
2 & $[0.009, 0.012[$ & $[0.103, 0.112[$ & $[0.15, 0.159[$ & $[0.226, 0.244[$ & $[0.066, 0.07[$ \\
3 & $[0.012, 0.014[$ & $[0.112, 0.121[$ & $[0.159, 0.165[$ & $[0.244, 0.256[$ & $[0.07, 0.073[$ \\
4 & $[0.014, 0.017[$ & $[0.121, 0.126[$ & $[0.165, 0.172[$ & $[0.256, 0.269[$ & $[0.073, 0.077[$ \\
5 & $[0.017, 0.02[$ & $[0.126, 0.131[$ & $[0.172, 0.179[$ & $[0.269, 0.282[$ & $[0.077, 0.08[$ \\
6 & $[0.02, 0.024[$ & $[0.131, 0.138[$ & $[0.179, 0.185[$ & $[0.282, 0.296[$ & $[0.08, 0.083[$ \\
7 & $[0.024, 0.03[$ & $[0.138, 0.143[$ & $[0.185, 0.193[$ & $[0.296, 0.31[$ & $[0.083, 0.089[$ \\
8 & $[0.03, 0.036[$ & $[0.143, 0.15[$ & $[0.193, 0.201[$ & $[0.31, 0.326[$ & $[0.089, 0.096[$ \\
9 & $[0.036, 0.048[$ & $[0.15, 0.162[$ & $[0.201, 0.215[$ & $[0.326, 0.36[$ & $[0.096, 0.106[$ \\
10 & $[0.048, 0.135[$ & $[0.162, 0.223[$ & $[0.215, 0.304[$ & $[0.36, 0.664[$ & $[0.106, 0.207[$ \\
\hline
\hline
\textbf{Bins} & \textbf{se fractal dimension} & \textbf{se symmetry} & \textbf{mean fractal dimension} & \textbf{se texture} & \textbf{se smoothness} \\
1 & $[0.001, 0.002[$ & $[0.008, 0.013[$ & $[0.05, 0.055[$ & $[0.36, 0.638[$ & $[0.002, 0.004[$ \\
2 & $[0.002, 0.002[$ & $[0.013, 0.015[$ & $[0.055, 0.057[$ & $[0.638, 0.78[$ & $[0.004, 0.005[$ \\
3 & $[0.002, 0.002[$ & $[0.015, 0.016[$ & $[0.057, 0.059[$ & $[0.78, 0.902[$ & $[0.005, 0.005[$ \\
4 & $[0.002, 0.003[$ & $[0.016, 0.017[$ & $[0.059, 0.06[$ & $[0.902, 1.005[$ & $[0.005, 0.006[$ \\
5 & $[0.003, 0.003[$ & $[0.017, 0.019[$ & $[0.06, 0.062[$ & $[1.005, 1.108[$ & $[0.006, 0.006[$ \\
6 & $[0.003, 0.004[$ & $[0.019, 0.02[$ & $[0.062, 0.063[$ & $[1.108, 1.239[$ & $[0.006, 0.007[$ \\
7 & $[0.004, 0.004[$ & $[0.02, 0.022[$ & $[0.063, 0.065[$ & $[1.239, 1.389[$ & $[0.007, 0.008[$ \\
8 & $[0.004, 0.005[$ & $[0.022, 0.026[$ & $[0.065, 0.068[$ & $[1.389, 1.562[$ & $[0.008, 0.009[$ \\
9 & $[0.005, 0.006[$ & $[0.026, 0.03[$ & $[0.068, 0.072[$ & $[1.562, 1.91[$ & $[0.009, 0.01[$ \\
10 & $[0.006, 0.03[$ & $[0.03, 0.079[$ & $[0.072, 0.097[$ & $[1.91, 4.885[$ & $[0.01, 0.031[$ \\
\end{tabular}
\end{adjustbox}
\caption{\textbf{Correspondence table between the bins resulting from the discretization process and the original features' values for the WDBC dataset.}}
\label{bin_wdbc}
\end{table}

\begin{table}
\centering
\small
\begin{adjustbox}{max width=\textwidth}
\begin{tabular}{*{6}{c}}
\hline
\textbf{Bins} & \textbf{hue} & \textbf{OD280$/$OD315} & \textbf{alcohol} & \textbf{color intensity} & \textbf{flavanoids} \\
 & & \textbf{of diluted wines} & & & \\
\hline
1 & $[1.28, 2.515[$ & $[278.0, 406.22[$ & $[11.03, 11.89[$ & $[0.48, 0.61[$ & $[0.34, 0.602[$ \\
2 & $[2.515, 2.902[$ & $[406.22, 472.32[$ & $[11.89, 12.25[$ & $[0.61, 0.74[$ & $[0.602, 0.843[$ \\ 
3 & $[2.902, 3.4[$ & $[472.32, 520.0[$ & $[12.25, 12.42[$ & $[0.74, 0.849[$ & $[0.843, 1.319[$ \\ 
4 & $[3.4, 4.068[$ & $[520.0, 604.76[$ & $[12.42, 12.754[$ & $[0.849, 0.91[$ & $[1.319, 1.731[$ \\ 
5 & $[4.068, 4.69[$ & $[604.76, 673.5[$ & $[12.754, 13.05[$ & $[0.91, 0.965[$ & $[1.731, 2.135[$ \\ 
6 & $[4.69, 5.286[$ & $[673.5, 743.2[$ & $[13.05, 13.283[$ & $[0.965, 1.04[$ & $[2.135, 2.466[$ \\ 
7 & $[5.286, 5.75[$ & $[743.2, 880.0[$ & $[13.283, 13.526[$ & $[1.04, 1.09[$ & $[2.466, 2.69[$ \\
8 & $[5.75, 7.044[$ & $[880.0, 1049.8[$ & $[13.526, 13.76[$ & $[1.09, 1.16[$ & $[2.69, 2.98[$ \\
9 & $[7.044, 8.578[$ & $[1049.8, 1263.9[$ & $[13.76, 14.1[$ & $[1.16, 1.238[$ & $[2.98, 3.238[$ \\
10 & $[8.578, 13.0[$ & $[1263.9, 1680.0[$ & $[14.1, 14.83[$ & $[1.238, 1.71[$ & $[3.238, 5.08[$ \\
\hline
\hline
\textbf{Bins} & \textbf{nonflavanoid phenols} & \textbf{malic acid} & \textbf{alcalinity of ash} & \textbf{proanthocyanins} & \textbf{ash} \\ 
\hline
1 & $[0.13, 0.212[$ & $[0.74, 1.242[$ & $[10.6, 16.0[$ & $[0.41, 0.844[$ & $[1.36, 2.0[$ \\
2 & $[0.212, 0.26[$ & $[1.242, 1.51[$ & $[16.0, 16.8[$ & $[0.844, 1.1[$ & $[2.0, 2.171[$ \\
3 & $[0.26, 0.28[$ & $[1.51, 1.65[$ & $[16.8, 17.972[$ & $[1.1, 1.336[$ & $[2.171, 2.259[$ \\
4 & $[0.28, 0.3[$ & $[1.65, 1.73[$ & $[17.972, 18.568[$ & $[1.336, 1.42[$ & $[2.259, 2.3[$ \\ 
5 & $[0.3, 0.34[$ & $[1.73, 1.865[$ & $[18.568, 19.5[$ & $[1.42, 1.555[$ & $[2.3, 2.36[$ \\
6 & $[0.34, 0.39[$ & $[1.865, 2.136[$ & $[19.5, 20.0[$ & $[1.555, 1.666[$ & $[2.36, 2.42[$ \\
7 & $[0.39, 0.43[$ & $[2.136, 2.691[$ & $[20.0, 21.0[$ & $[1.666, 1.87[$ & $[2.42, 2.5[$ \\
8 & $[0.43, 0.48[$ & $[2.691, 3.428[$ & $[21.0, 22.0[$ & $[1.87, 1.99[$ & $[2.5, 2.61[$ \\
9 & $[0.48, 0.53[$ & $[3.428, 3.988[$ & $[22.0, 24.0[$ & $[1.99, 2.329[$ & $[2.61, 2.7[$ \\
10 & $[0.53, 0.66[$ & $[3.988, 5.8[$ & $[24.0, 30.0[$ & $[2.329, 3.58[$ & $[2.7, 3.23[$ \\
\hline
\hline
\textbf{Bins} & \textbf{proline} & \textbf{total phenols} & \textbf{magnesium} \\
\hline
1 & $[1.27, 1.58[$ & $[0.98, 1.457[$ & $[70.0, 85.0[$ \\
2 & $[1.58, 1.781[$ & $[1.457, 1.65[$ & $[85.0, 88.0[$ \\
3 & $[1.781, 2.137[$ & $[1.65, 1.876[$ & $[88.0, 90.0[$ \\
4 & $[2.137, 2.517[$ & $[1.876, 2.05[$ & $[90.0, 94.68[$ \\
5 & $[2.517, 2.78[$ & $[2.05, 2.355[$ & $[94.68, 98.0[$ \\
6 & $[2.78, 2.903[$ & $[2.355, 2.53[$ & $[98.0, 101.0[$ \\
7 & $[2.903, 3.12[$ & $[2.53, 2.7[$ & $[101.0, 105.0[$ \\
8 & $[3.12, 3.26[$ & $[2.7, 2.86[$ & $[105.0, 111.0[$ \\
9 & $[3.26, 3.466$ & $[2.86, 3.082[$ & $[111.0, 118.0[$ \\
10 & $[3.466, 4.0[$ & $[3.082, 3.88[$ & $[118.0, 162.0[$ \\
\hline
\end{tabular}
\end{adjustbox}
\caption{\textbf{Correspondence table between the bins resulting from the discretization process and the the original features' values for the Wine dataset.}}
\label{bin_wine}
\end{table}

\begin{table}
\centering
\small
\begin{adjustbox}{max width=\textwidth}
\begin{tabular}{ccccc}
\hline
\textbf{Bins} & \textbf{sepal width} & \textbf{petal width} & \textbf{sepal length} & \textbf{petal length} \\
\hline
1 & $[2.0, 2.5[$ & $[0.1, 0.2[$ & $[4.3, 4.8[$ & $[1.0, 1.4[$ \\
2 & $[2.5, 2.7[$ & $[0.2, 0.2[$ & $[4.8, 5.0[$ & $[1.4, 1.5[$ \\
3 & $[2.7, 2.8[$ & $[0.2, 0.4[$ & $[5.0, 5.246[$ & $[1.5, 1.7[$ \\
4 & $[2.8, 3.0[$ & $[0.4, 1.148[$ & $[5.246, 5.6[$ & $[1.7, 3.9[$ \\
5 & $[3.0, 3.0[$ & $[1.148, 1.3[$ & $[5.6, 5.8[$ & $[3.9, 4.35[$ \\
6 & $[3.0, 3.1[$ & $[1.3, 1.5[$ & $[5.8, 6.1[$ & $[4.35, 4.652[$ \\
7 & $[3.1, 3.2[$ & $[1.5, 1.8[$ & $[6.1, 6.3[$ & $[4.652, 5.0[$ \\
8 & $[3.2, 3.4[$ & $[1.8, 1.9[$ & $[6.3, 6.556[$ & $[5.0, 5.356[$ \\
9 & $[3.4, 3.658[$ & $[1.9, 2.2[$ & $[6.556, 6.9[$ & $[5.356, 5.8[$ \\
10 & $[3.658, 4.4[$ & $[2.2, 2.5[$ & $[6.9, 7.9[$ & $[5.8, 6.9[$ \\
\hline
\end{tabular}
\end{adjustbox}
\caption{\textbf{Correspondence table between the bins resulting from the discretization process and the original features' values for the Iris dataset.}}
\label{bin_iris}
\end{table}

\begin{table}
\centering
\small
\begin{adjustbox}{max width=\textwidth}
\begin{tabular}{*{7}{c}}
\hline
\textbf{Bins} & \textbf{ca} & \textbf{oldpeak} & \textbf{thalach} & \textbf{age} & \textbf{chol} & \textbf{trestbps} \\
\hline
1 & $[0.0, 0.0[$ & $[0.0, 0.0[$ & $[71.0, 115.72[$ & $[29.0, 42.0[$ & $[126.0, 188.0[$ & $[94.0, 110.0[$ \\
2 & $[0.0, 0.0[$ & $[0.0, 0.0[$ & $[115.72, 130.0[$ & $[42.0, 45.0[$ & $[188.0, 204.0[$ & $[110.0, 120.0[$ \\
3 & $[0.0, 0.0[$ & $[0.0, 0.0[$ & $[130.0, 140.36[$ & $[45.0, 50.0[$ & $[204.0, 218.0[$ & $[120.0, 120.0[$ \\
4 & $[0.0, 0.0[$ & $[0.0, 0.368[$ & $[140.36, 146.0[$ & $[50.0, 53.0[$ & $[218.0, 230.0[$ & $[120.0, 126.0[$ \\
5 & $[0.0, 0.0[$ & $[0.368, 0.8[$ & $[146.0, 153.0[$ & $[53.0, 56.0[$ & $[230.0, 241.0[$ & $[126.0, 130.0[$ \\
6 & $[0.0, 1.0[$ & $[0.8, 1.132[$ & $[153.0, 159.0[$ & $[56.0, 58.0[$ & $[241.0, 254.0[$ & $[130.0, 134.0[$ \\
7 & $[1.0, 1.0[$ & $[1.132, 1.4[$ & $[159.0, 163.0[$ & $[58.0, 59.64[$ & $[254.0, 268.64[$ & $[134.0, 140.0[$ \\
8 & $[1.0, 1.0[$ & $[1.4, 1.9[$ & $[163.0,170.0[$ & $[59.64, 62.0[$ & $[268.64, 286.0[$ & $[140.0, 144.96[$ \\
9 & $[1.0, 2.0[$ & $[1.9, 2.8[$ & $[170.0, 177.28[$ & $[62.0, 66.0[$ & $[286.0, 309.0[$ & $[144.96, 152.56[$ \\
10 & $[2.0, 3.0[$ & $[2.8, 6.2[$ & $[177.28, 202.0[$ & $[66.0, 77.0[$ & $[309.0, 564.0[$ & $[152.56, 200.0[$ \\
\hline
\hline
\textbf{Bins} & \textbf{thal} & \textbf{cp} & \textbf{exang} & \textbf{slope} & \textbf{sex} & \textbf{restecg} \\
\hline
1 & normal & typical angina & no & up-sloping & female & normal \\
2 & fixed defect & atypical & yes & flat & male & ST-T wawe \\
 &  & angina & & & & abnormality \\
3 & reversed defect & non-anginal & - & down-sloping & - & left ventricular \\
 & & pain & & & & abnormality \\
4 & - & asymptomatic & - & - & - & -  \\
\hline
\hline
\textbf{Bins} & \textbf{fbs} &  &  &  & \\
\hline
1 & no &  &  &  &  &  \\
2 & yes &  &  &  &  & \\
\end{tabular}
\end{adjustbox}
\caption{\textbf{Correspondence table between the bins resulting from the discretization process and the original features' values for the Heart Disease dataset.}}
\label{bin_heart}
\end{table}

\begin{table}
\centering
\small
\begin{adjustbox}{max width=\textwidth}
\begin{tabular}{*{5}{c}}
\hline
\textbf{Bins} & \textbf{x1} & \textbf{x2} & \textbf{x3} & \textbf{x4} \\
\hline
1 & $[0.003, 0.102[$ & $[0.001, 0.087[$ & 0 & blue \\
2 & $[0.102, 0.208[$ & $[0.087, 0.175[$ & 1 & white \\
3 & $[0.208, 0.289[$ & $[0.175, 0.293[$ & - & red \\
4 & $[0.289, 0.408[$ & $[0.293, 0.384[$ & - & - \\
5 & $[0.408, 0.53[$ & $[0.384, 0.473[$ & - & - \\
6 & $[0.53, 0.632[$ & $[0.473, 0.583[$ & - & - \\
7 & $[0.632, 0.715[$ & $[0.583, 0.689[$ & - & - \\
8 & $[0.715, 0.788[$ & $[0.689, 0.796[$ & - & - \\
9 & $[0.788, 0.905[$ & $[0.796, 0.891[$ & - & - \\
10 & $[0.905, 0.999[$ & $[0.891, 0.998[$ & - & - \\
\hline
\end{tabular}
\end{adjustbox}
\caption{\textbf{Correspondence table between the bins resulting from the discretization process and the original features' values for the Synthetic and Synthetic noisy datasets.}}
\label{bin_synthetic}
\end{table}

\newpage
\vskip 0.2in
\bibliography{sample}

\end{document}